\documentclass[lettersize,journal]{IEEEtran}
\usepackage{graphicx}
\usepackage{cite}
\usepackage{picinpar}
\usepackage{amsmath}
\usepackage{url}
\usepackage{flushend}
\usepackage[latin1]{inputenc}
\usepackage{colortbl}
\usepackage{soul}
\usepackage{multirow}
\usepackage{pifont}
\usepackage{color}
\usepackage{alltt}
\usepackage[hidelinks]{hyperref}
\usepackage{enumerate}
\usepackage{siunitx}
\usepackage{breakurl}
\usepackage{epstopdf}
\usepackage{pbox}
\usepackage{tabularx}
\usepackage{booktabs}

\usepackage{graphics} % for pdf, bitmapped graphics files
\usepackage{graphicx}
\usepackage{epsfig} % for postscript graphics files
\usepackage{times} % assumes new font selection scheme installed
\usepackage{amsmath} % assumes amsmath package installed
\usepackage{amssymb}  % assumes amsmath package installed

\usepackage{bm}
\usepackage[linesnumbered,ruled,vlined]{algorithm2e}
\usepackage[noend]{algpseudocode}
\usepackage[font={small}]{caption}
\usepackage{subcaption}
\usepackage{physics}
\usepackage{xcolor}
\usepackage{tikz}
\usetikzlibrary {arrows.meta}
\usetikzlibrary{fit,
                positioning}
\usepackage[belowskip=1ex]{subcaption}  % need to be added
\usetikzlibrary{calc} 
\usetikzlibrary{positioning}
\usepackage{balance}
\usepackage{soul}
\usetikzlibrary{arrows}
\usepackage{cite}
\usepackage{url}
\usepackage{multirow}
\usepackage{booktabs} 

\usepackage{setspace}
\newcommand{\mexr}[1]{${#1}$\xspace}

\newcommand{\mn}[2]{{#1}_\textnormal{#2}}
\newcommand{\tn}[2]{\mexr{\mn{#1}{#2}}}
\newcommand{\mb}[2]{\mathbf{#1}_\textnormal{#2}}
\newcommand{\tb}[2]{\mexr{\mb{#1}{#2}}}
\newcommand{\pdfm}[1]{\mn{\Phi}{#1}}

\newcommand{\X}[1]{\tn{X}{#1}}
\newcommand{\Xm}[1]{\mn{X}{#1}}

\newcommand{\x}[1]{\tb{x}{#1}}
\newcommand{\xm}[1]{\mb{x}{#1}}

\newcommand{\sminm}{s_\textnormal{min}}
\newcommand{\smin}{\mexr{\sminm}}

\newcommand{\ahism}{\hat{\sigma}_\textnormal{adms}}
\newcommand{\ahis}{\mexr{\ahism}}

\newcommand{\cahism}{\hat{s}_\textnormal{adms}}
\newcommand{\cahis}{\mexr{\cahism}}

\newcommand{\ehism}{\Tilde{\sigma}_\textnormal{est}}
\newcommand{\ehis}{\mexr{\ehism}}

\newcommand{\cehism}{\Tilde{s}_\textnormal{est}}
\newcommand{\cehis}{\mexr{\cehism}}

\newcommand{\eism}{X_{\Tilde{\sigma}}}
\newcommand{\eis}{\mexr{\eism}}

\begin{document}
\title{Estimated Informed Anytime Search for Sampling-Based Planning via Adaptive Sampler}
%%%@@@@@@@@@@@@@@@@@@@@@@@
\author{
	\vskip 1em
	
	Liding Zhang$^{1}$, Kuanqi Cai$^{2}$, Yu Zhang$^{1}$, Zhenshan Bing$^{3,1}$, Chaoqun Wang$^{4}$, Fan Wu$^{2}$,\\ Sami Haddadin$^{2}$, \emph{Fellow, IEEE}, and Alois Knoll$^{1}$, \emph{Fellow, IEEE}

    % First A. Author, \IEEEmembership{Fellow, IEEE}, Second B. Author, and Third C. Author, Jr., \IEEEmembership{Member, IEEE}

	\thanks{
	
		% Manuscript received Month xx, 2xxx; revised Month xx, xxxx; accepted Month x, xxxx.
		% This work was supported in part by the xxx Department of xxx under Grant  (sponsor and financial support acknowledgment goes here).
		
		% (Authors' names and affiliation) First A. Author1 and Second B. Author2 are with the xxx Department, University of xxx, City, Zip code, Country, on leave from the National Institute for xxx, City, Zip code, Country (e-mail: author@domain.com). 
		
		% Third C. Author3 is with the National Institute of xxx, City, Zip code, Country (corresponding author to provide phone: xxx-xxx-xxxx; fax: xxx-xxx-xxxx; e-mail: author@ domain.gov).
        $^{1}$L. Zhang, Y. Zhang, Z. Bing, and A. Knoll are  with the Chair of Robotics, Artificial Intelligence and Real-time Systems, TUM School of Computation, Information and Technology (CIT), Technical University of Munich, 85748 Garching bei Munich, Germany.
        {\tt\small liding.zhang@tum.de}%
	}
 
         \thanks{$^{2}$K. Cai, F. Wu, S. Haddadin is with the Chair of Robotics and Systems Intelligence, Munich Institute of Robotics and Machine Intelligence (MIRMI), Technical University of Munich, 80992 Munich, Germany.}
         \thanks{$^{3}$Zhenshan Bing is also with the State Key Laboratory for Novel Software Technology and the School of Science and Technology, Nanjing University (Suzhou Campus), China.}
         \thanks{$^{4}$C. Wang is with the School of Control Science and Engineering, Shandong University, 250100 Shandong, China. \\
        \textit{(Corresponding authors: Zhenshan Bing; Kuanqi Cai.)}}
 
}

\maketitle

\begin{abstract}
Path planning in robotics often involves solving continuously valued, high-dimensional problems. Popular informed approaches include graph-based searches, such as A*, and sampling-based methods, such as Informed RRT*, which utilize informed set and anytime strategies to expedite path optimization incrementally. Informed sampling-based planners define informed sets as subsets of the problem domain based on the current best solution cost. However, when no solution is found, these planners re-sample and explore the entire configuration space, which is time-consuming and computationally expensive. This article introduces Multi-Informed Trees (MIT*), a novel planner that constructs estimated informed sets based on prior admissible solution costs before finding the initial solution, thereby accelerating the initial convergence rate. Moreover, MIT* employs an adaptive sampler that dynamically adjusts the sampling strategy based on the exploration process. Furthermore, MIT* utilizes length-related adaptive sparse collision checks to guide lazy reverse search. These features enhance path cost efficiency and computation times while ensuring high success rates in confined scenarios. Through a series of simulations and real-world experiments, it is confirmed that MIT* outperforms existing single-query, sampling-based planners for problems in $\mathbb{R}^4$ to $\mathbb{R}^{16}$ and has been successfully applied to real-world robot manipulation tasks. A video showcasing our experimental results is available at: \href{https://youtu.be/30RsBIdexTU}{\textcolor{blue}{https://youtu.be/30RsBIdexTU}}.

\emph{Note to Practitioners---} The motivation for this work stems from the challenges faced by existing informed path planners in high-dimensional, continuously valued environments, particularly when an initial feasible solution is difficult to find. \textcolor{black}{Traditional asymmetric bidirectional planners rely on the best current solution to define problem subsets. When a lazy path has been found through lazy reverse search, these planners tend to re-sample and explore the entire problem space, which could hinder the path planning process.}
Our proposed MIT* algorithm addresses this issue by constructing an estimated informed set based on prior admissible solution costs before finding the initial solution. This estimated set helps to narrow the search area, thereby accelerating the initial convergence rate. MIT* also integrates an adaptive sampling strategy that dynamically adjusts based on the ongoing exploration process, enhancing the planner's ability to efficiently navigate through challenging spaces. Furthermore, MIT* employs adaptive sparse collision checks, which guide the lazy reverse search that balances computational efficiency with accuracy in pathfinding.
The proposed algorithm can be applied to industrial robots, humanoid robots, or service robots to achieve efficient path planning.
\end{abstract}


\markboth{IEEE Transactions on Automation Science and Engineering}%
{}

\definecolor{limegreen}{rgb}{0.2, 0.8, 0.2}
\definecolor{forestgreen}{rgb}{0.13, 0.55, 0.13}
\definecolor{greenhtml}{rgb}{0.0, 0.5, 0.0}

\section{Introduction}
\IEEEPARstart{T}{rajectory} planning is a fundamental challenge in robotics, aiming to devise a feasible path for a robot while avoiding collisions. This task can be framed as finding collision-free paths from a start to a goal state in state space while avoiding obstacles~\cite{Gammell2021,zhang2024review}. In practical applications, path planning for robot manipulators remains computationally intensive, particularly in high-dimensional spaces. The most resource-demanding aspect of this process is often the collision checking step. To address these challenges, a variety of methods have been developed, each designed to enhance the efficiency and effectiveness of robot path planning.

Popular graph-based algorithms, such as Dijkstra's~\cite{dijkstra1959note}, find the shortest path by exploring all routes in a discrete graph, while the A* algorithm~\cite{hart1968formal} improves efficiency with heuristic search. The Anytime Repairing A* (ARA*)~\cite{likhachev2003ara} provides progressively optimal solutions, ensuring feasible paths at \textit{anytime}. However, applying these methods to continuous spaces requires discretization, which is either sparse (yielding suboptimal paths) or dense (incurring high computational costs due to the \textit{curse of dimensionality}~\cite{bellman1957dynamic}).  
\textcolor{black}{To tackle this trade-off, sampling-based planners such as Rapidly-exploring Random Trees (RRT)~\cite{LaValle1998}, Expansive Space Trees (EST)~\cite{Hsu2002}, and Probabilistic Roadmaps (PRM)~\cite{kavraki1998analysis} operate directly in the configuration space (\textit{$\mathcal{C}$-space}). By sampling configurations and connecting them via local planners with collision checking, these methods efficiently solve high-dimensional planning problems without prior discretization of \( \mathcal{C}\)-space~\cite{zhang2025apt}.}

RRT-Connect~\cite{kuffner2000rrt} extends RRT to efficiently find a path between two points in \textit{$\mathcal{C}$-space} by utilizing two RRTs to connect start and goal states~\cite{zhang2025G3t}. RRT*~\cite{karaman2011sampling} builds on RRT by incrementally rewiring the tree to ensure asymptotic optimality. \textcolor{black}{In the context of lazy collision checking~\cite{strub2022adaptively}, Lazy PRM~\cite{bohlin2000path} and Lazy PRM*~\cite{hauser2015lazy} assume all edges are initially valid and delay the expensive collision checks until after a path is found.} This postponement of collision checks accelerates the planning process. To narrow the search domain, Informed RRT*~\cite{gammell2014informed} refines RRT* by employing elliptical informed sampling rely on the current best solution cost (i.e., the \textit{informed set})~\cite{gammell2014informed,gammell2018informed}, which simplifies the search and sampling set, thereby enhancing the convergence rate. The \textit{greedy informed set}~\cite{Phone2022greedyinformed} further improves convergence to the optimal solution by basing the informed set on the maximum admissible estimated cost of the current path. \textcolor{black}{Guided Incremental Local Densification (GuILD)~\cite{Scalise2023} enhances path optimization by defining local subsets of problem domain via beacon selector. However, these sets can only be constructed when an initial solution has been found~\cite{zhang2025fuzzy}. When no solution is available, the planners inefficiently re-sample and explore the entire \textit{$\mathcal{C}$-space}~\cite{gammell2018informed}.}

% \subsection{Proposed Algorithm and Original Contributions}
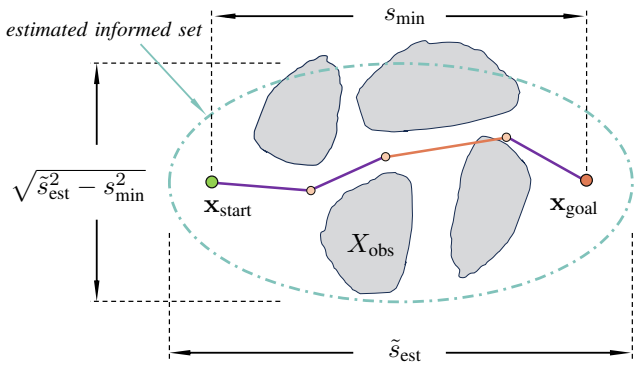
\begin{figure}[t!]
    \centering
    \begin{tikzpicture}
   
    \node[inner sep=0pt] (russell) at (0.0,0.0)
    {\includegraphics[width=0.4\textwidth]{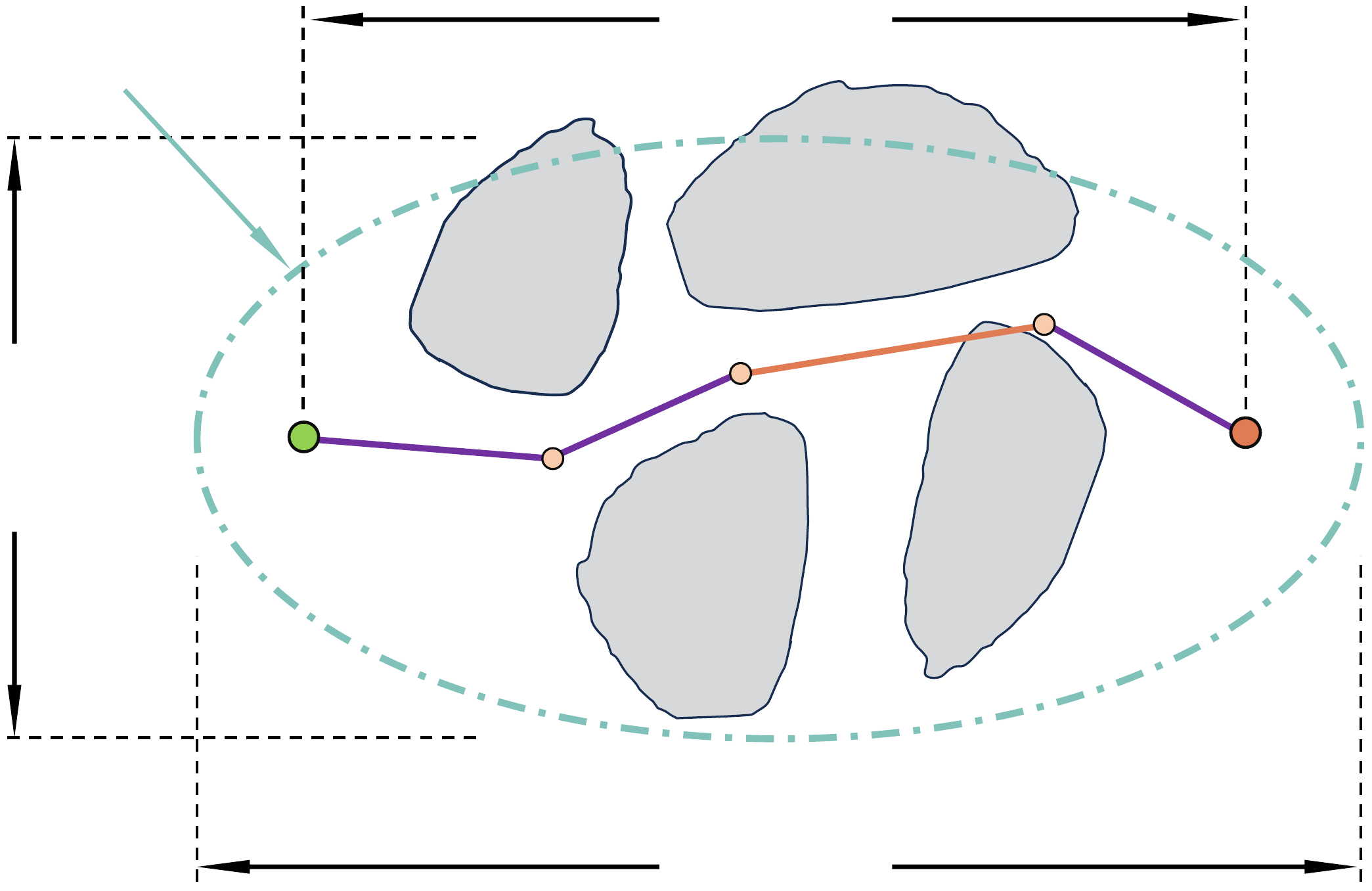}};
    \node at (0.55,2.25) {\smin};
    \node at (0.55,-2.2) {\cehis};
    \node at (-1.8,-0.3) {$\mathbf{x}_\textnormal{start}$};
    \node at (2.8,-0.3) {$\mathbf{x}_\textnormal{goal}$};
    % \node at (0.1,-0.8) {$X_\textnormal{obs}$};
    \node at (-3.8,0.0) {$\sqrt{\cehism^2 - \sminm^2}$};
    \node at (-3.45,2.05) {\footnotesize\textit{estimated informed set}};
    \node at (0.1,-0.8) {\X{obs}};
    
    % \node at (0.55,-0.88) {$\textit{informed set}$};
        
    \end{tikzpicture}
    % \vspace{-0.5em} 
    \caption{Illustration of the $L^2$ estimated informed set, it's determined by the start, goal states, and two key costs: the theoretical minimum cost $s_\textnormal{min}$ and the current expand \textit{estimate initial solution} cost $\cehism$.}
    \label{fig: elipseEIS}
    \vspace{-1.7em} 
\end{figure}
In this article, we propose a novel sampling-based planner, Multi-Informed Trees (MIT*), which is multi-informed by the prior admissible cost during the initial pathfinding and the current best cost during the optimization. The prior admissible cost is calculated from the lazy reverse search to construct an \textit{estimated informed set} (Fig.~\ref{fig: elipseEIS}). By combining prior costs with a reliability expansion factor, MIT* explores within an ellipsoidal subset of the planning domain, thereby guiding the search while no initial solution has been established yet.
Furthermore, MIT* employs an adaptive sampler that adjusts its strategy based on the exploration process, which densifies distribution around obstacles. MIT* improves reverse search by adjusting resolution based on edge length to enhance computational efficiency while maintaining search effectiveness.

\textcolor{black}{MIT* outperforms state-of-the-art (SOTA) methods in initial solution time, initial solution cost (e.g., path length), and final solution cost (i.e., optimized path) across generalized simulation benchmarks and various real-world experiments.}

The contributions of this work are summarized as follows:
\begin{enumerate}
\item \textit{Estimated informed set:} Constructs a set before finding the initial path, using prior costs and expansion factors to effectively guide the exploration and sampling process.
\item \textit{Adaptive sampling strategy:} Integrates dynamic sampling adjustments to improve distribution around obstacles and in narrow corridors based on the exploration process.
\item \textit{Real-world application in robotic manipulation:} MIT* was evaluated in versatile real-world tasks and achieved the highest success rate across multiple trials.
\end{enumerate}

The rest of this article is organized as follows. Section~\ref{sec:prob} introduces the problem definition. In Section~\ref{sec:method}, the MIT* algorithm with the proof of asymptotic optimality, is presented in detail. Simulation and experimental results are discussed in Section~\ref{sec:Expri}. Finally, Section~\ref{sec:conclu} concludes this article.

\section{Related Work}
In sampling-based motion planning, a common drawback is the slow convergence to an optimal solution. This section introduces popular methods to expedite path optimization.
\subsection{Informed Optimal Path Planning}
Recently, researchers have proposed various heuristics to improve the convergence of sampling-based algorithms to an optimal path~\cite{zhang2025dit, wang2023, ZHANG2025git}. Batch Informed Trees (BIT*)~\cite{gammell2020batch} builds on Informed RRT* by constructing an implicit \textit{random geometric graph} (RGG)~\cite{penrose2003random} and performing a step-wise search similar to Lifelong Planning A* (LPA*)~\cite{koenig2004lifelong}. BIT* focuses on almost-surely asymptotic optimality by refining the RGG through batch processing~\cite{Zhang2024adaptive}, aggregating multiple state batches to form a denser graph over time, which reduces computational overhead while improving solution quality. Advanced BIT* (ABIT*)~\cite{strub2020advanced} introduces inflation and truncation factors to balance exploration and exploitation in the denser RGG approximation, enhancing its effectiveness in complex environments. \textcolor{black}{Energy-efficient BIT* for
reconfigurable robots (EBITR*)~\cite{Phone2022greedyinformed} utilizes a greedy informed set, based solely on the maximum heuristic cost of the current solution's state. Adaptively Informed Trees (AIT*)~\cite{strub2022adaptively, strub2020adaptively} use adaptive heuristics and an asymmetrical search strategy, incorporating sparse collision checks during the reverse search phase to improve precision by continuously updating the heuristic based on the evolving approximation. BiAIT*~\cite{Li2024} employs symmetrical bidirectional search for both heuristic and space searching. Effort Informed Trees (EIT*)~\cite{strub2022adaptively}, a SOTA planner, utilize admissible cost (i.e., the lower bound on the true value) and effort heuristics (e.g., \textit{the number of collision checks}) to optimize objectives with obstacle clearance~\cite{Zhang2024Elliptical}.}

Despite these advancements, these planners typically rely on uniform sampling strategies, which may struggle with critical zones such as narrow passages. Uniform sampling often has a lower probability of sampling in such challenging regions, potentially impacting the efficiency and effectiveness of pathfinding in complex, high-dimensional spaces.
\textcolor{black}{\subsection{Non-uniform Sampling Methods}}
\textcolor{black}{Improved sampling strategies are crucial for navigating complex environments. Learning-based sampling~\cite{Ichter2018} distribution from demonstrations using a conditional variational autoencoder, Neural RRT*~\cite{Wang2020} trains a neural network to perform nonuniform sampling, leading to more efficient tree expansion.} Techniques like Obstacle-based PRM (OBPRM)~\cite{amato1998obprm} enhance sampling distribution near obstacles by generating samples close to obstacle surfaces through perturbation, which improves path discovery in constrained regions. Gaussian sampling~\cite{boor1999gaussian} biases samples toward obstacle boundaries, increasing the likelihood of finding paths through tight spaces by placing one sample inside and another outside an obstacle's influence.
Bridge test sampling~\cite{hsu2003bridge} connects pairs of samples with a \textit{bridge} to detect narrow passages, concentrating sampling efforts on critical pathways in cluttered environments. However, OBPRM may result in uneven sample distribution, Gaussian sampling assumes that data follows a normal distribution, and Bridge Test samplers focus only on the midpoint, potentially missing key samples along the edge.

In contrast to previous work, we utilize prior knowledge from lazy search attempts (i.e., prior admissible costs) to define the estimated informed set. Moreover, we implement an adaptive sampler that adjusts its sampling strategy dynamically and increases sampling density in critical zones.
\vspace{0.2cm}
\section{Preliminaries}\label{sec:prob}
\textcolor{black}{In this section, we first formulate the optimal path planning and then provide notations for the MIT* algorithm (Alg.~\ref{alg: mit}).}
\subsection{Problem Formulation}
\textit{Definition 1 (Optimal Planning):} Define a planning problem with the state space $X \subseteq \mathbb{R}^n$. Let $X_{\textnormal{obs}} \subset X$ represent states in collision with obstacles, and $X_{\textnormal{free}} = \textnormal{cl}(X \setminus X_{\textnormal{obs}})$ denote the resulting permissible states, where $\textnormal{cl}(\cdot)$ represents the \textit{closure} of a set. The initial state is denoted by $\mathbf{x}_{\textnormal{start}} \in X_{\textnormal{free}}$, and the set of desired final states is $X_{\textnormal{goal}} \subset X_{\textnormal{free}}$. A sequence of states $\sigma: [0, 1] \mapsto X$ forms a continuous map (i.e., a collision-free path), and $\Sigma$ represents the set of all nontrivial paths~\cite{karaman2011sampling}.

\textcolor{black}{The optimal solution, represented as $\sigma^*$, corresponds to the path that minimizes a selected cost function $c: \Sigma \mapsto \mathbb{R}_{\geq 0}$. This path connects the initial state $\mathbf{x}_{\textnormal{start}}$ to any goal state $\mathbf{x}_{\textnormal{goal}} \in X_{\textnormal{goal}}$ through the free space:
\begin{equation}
\begin{split}
    \sigma^* &= \arg \min_{\sigma \in \Sigma} \left\{ c(\sigma) \,\middle|\, \sigma(0) = \mathbf{x}_{\textnormal{start}}, \sigma(1) \in \mathbf{x}_{\textnormal{goal}}, \right. \\
    &\qquad\qquad \left. \forall t \in [0, 1], \sigma(t) \in X_{\textnormal{free}} \right\}.
\end{split}
\end{equation}
where $\mathbb{R}_{\geq 0}$ denotes the non-negative real numbers. The cost of the optimal path is $c^*$, and \( t \) represents the continuous time parameter over the path $\sigma$.
Considering set of states, $X_{\textnormal{samples}} \subset X$, as a graph where a transition function determines edges, we can describe its properties using a probabilistic model implicit in dense RGGs when these states are randomly sampled, i.e., $X_{\textnormal{samples}} = \{ \mathbf{x} \sim X_\textnormal{free} \}$, as discussed in~\cite{penrose2003random}.}

\textit{Definition 2 (Potential subsets of problem domain):} Let \( g(\xm{}) \) denote the cost of the optimal path from the start to a state \( \xm{} \in X_{\textnormal{free}} \), where the optimal \textit{cost-to-come} is defined as:
\begin{equation}
g(\xm{}) := \min_{\sigma \in \Sigma} \{c(\sigma) \mid \sigma(0) = \xm{start}, \sigma(1) = \xm{}\},
\end{equation}
Let \( h(\xm{}) \) denote the cost of the optimal path from \( \xm{} \) to the goal region, where the optimal \textit{cost-to-go} is defined as:
\begin{equation}
h(\xm{}) := \min_{\sigma \in \Sigma} \{c(\sigma) \mid \sigma(0) = \xm{}, \sigma(1) \in X_{\textnormal{goal}}\},
\end{equation}
\textcolor{black}{The cost of the optimal path from \( \xm{start} \) to \( X_{\textnormal{goal}} \), constrained to pass through \( \xm{} \), is given by \( f(\xm{}) := g(\xm{}) + h(\xm{}) \). This defines the states within the subsets that can potentially yield a solution better than the current solution \( c_i \) as:
\begin{equation}
\Xm{subset} := \{ \xm{} \in X_{\textnormal{free}} \mid f(\xm{}) < c_i \}.
\end{equation}
The defined set \( \Xm{subset} \subseteq X \) represents a subset of the problem domain and depends on the cost (Fig.~\ref{fig: elipseEIS}), as discussed in~\cite{gammell2018informed}.}
\subsection{Notation}~\label{subsec: notation}
The state space is defined as $X \subseteq \mathbb{R}^n$ with the initial state $\mathbf{x}_{\textnormal{init}} \in X$ and target states $X_{\textnormal{goal}} \subset X$. Sampled states are noted as $X_{\textnormal{sampled}}$. Forward and reverse trees are represented as $\mathcal{T_F} = (V_\mathcal{F}, E_\mathcal{F})$ and $\mathcal{T_R} = (V_\mathcal{R}, E_\mathcal{R})$, respectively. The nodes $V_\mathcal{F}$ and $V_\mathcal{R}$ correspond to valid states. Edges $E_\mathcal{F} \subset V_\mathcal{F} \times V_\mathcal{F}$ in the forward tree link validly connected states, while edges $E_\mathcal{R} \subset V_\mathcal{R} \times V_\mathcal{R}$ in the reverse tree may traverse invalid areas. Each edge $(\mathbf{x}_s, \mathbf{x}_t)$ connects source state $\mathbf{x}_s$ to target state $\mathbf{x}_t$.

The function $c: X \times X \rightarrow [0, \infty)$ denotes the actual cost (path length) of connecting two states, and $\hat{c}: X \times X \rightarrow [0, \infty)$ provides an approximation ensuring $\hat{c}(\mathbf{x}_i, \mathbf{x}_j) \leq c(\mathbf{x}_i, \mathbf{x}_j)$. The estimated costs from the initial state to any state are given by $\hat{g}(\mathbf{x}):= \hat{c}(\mathbf{x}_{\textnormal{start}}, \mathbf{x})$. The admissible cost heuristics to reach the goal are denoted by $\hat{h}(\mathbf{x}):= \min_{\mathbf{x}_{\textnormal{goal}} \in X_{\textnormal{goal}}} \hat{c}(\mathbf{x}, \mathbf{x}_{\textnormal{goal}})$.
The forward tree cost from the initial state to a given state is \( g_\mathcal{F}: X \rightarrow [0, \infty) \). The path cost from start to goal via a state is \( \hat{f}(\mathbf{x}) := \hat{g}(\mathbf{x}) + \hat{h}(\mathbf{x}) \). The informed set \( X_{\hat{f}} := \{\mathbf{x} \in X | \hat{f}(\mathbf{x}) < c_\textnormal{curr} \} \) contains states that can improve the cost of the current solution, denotes as $c_\textnormal{curr}$.

Let \( A \) be a set and \( B, C \) be subsets of \( A \). The notation \( B \stackrel{+}{\leftarrow} C \) denotes \( B \leftarrow B \cup C \) and \( B \stackrel{-}{\leftarrow} C \) denotes \( B \leftarrow B \setminus C \).

\textit{MIT* specific notation:}
The admissible initial solution is denoted by \ahis and its cost by $\cahism: X \rightarrow [0, \infty)$. The estimated initial solution is denoted by \ehis and its cost by $\cehism$. The estimated informed set is denoted as $\eism$. The reliability parameter of $\cahism$ is $\gamma \in (0,1)$. The expand factor of $\eism$ is $e_{\gamma}$. The number of sparse checks per edge in iteration $k$ is $\Theta_\textnormal{sparse,$k$}$. The resolution of sparse checks for edges is $\Delta_\textnormal{sparse,$k$}$.
\section{Multi-Informed Trees (MIT*)}\label{sec:method}
\begin{figure*}[t!] % [h]选项尝试将图片放置在当前位置
    \centering % 让图片居中显示
    \includegraphics[width=0.9\textwidth]{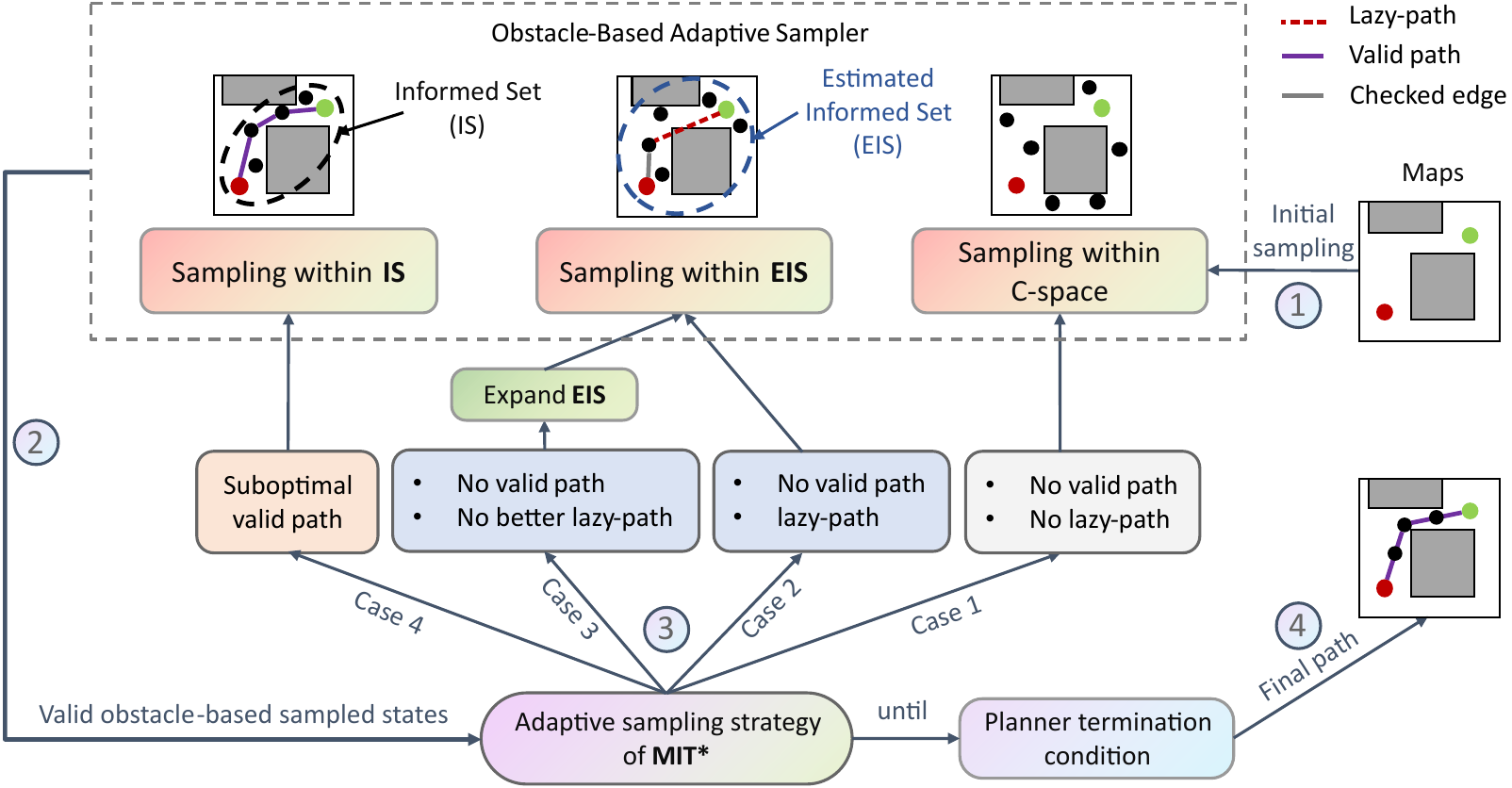} % 调整图片大小并插入图片
    \caption{The path planning process of MIT* begins with initial sampling in the entire \textit{$\mathcal{C}$-space} and selecting an adaptive sampler case. If no initial path/lazy path is found, re-sampling occurs in the entire \textit{$\mathcal{C}$-space} (i.e., case 1). If a lazy path is found but invalid, sampling is conducted within the estimated informed set, expanding it when the lazy path cannot be improved (i.e., cases 2, 3). Upon finding a valid path, the informed set is calculated and optimized until the planner's termination condition is met (i.e., case 4).} % 添加图片标题
    \label{fig:generalConcept} % 为图片添加标签，方便在文中引用
    \vspace{-1.4em} 
\end{figure*}

In this section, we first introduce the concept of obstacle-based adaptive sampling. Next, we illustrate the estimated informed set. Then, we propose a sparse collision check based on the edge length. Finally, we analyze the probabilistic completeness and asymptotic optimality of MIT*.
\begin{algorithm}[t]
\caption{Multi-Informed Trees (MIT*)}
\SetKwInOut{Input}{Input}
\SetKwInOut{Output}{Output}
\SetKwFunction{sample}{sample}
\SetKwFunction{adaptiveSampler}{adaptiveSampler}
\SetKwFunction{updateEIS}{updateEIS}
\SetKwFunction{bestKey}{bestKey}
\SetKwFunction{lazyReverseSearch}{lazyReverseSearch}
\SetKwFunction{updateLazyReverseSearch}{updateLazyReverseSearch}
\SetKwFunction{forwardSearch}{forwardSearch}
\SetKwFunction{couldImproveForwardSearch}{couldImproveForwardSearch}
\SetKwFunction{isCollide}{isCollide}
\SetKwFunction{lazyCheck}{lazyCheck}
\SetKwFunction{terminateCondition}{terminateCondition}
\SetKwFunction{prune}{prune}

\DontPrintSemicolon
\small
\label{alg: mit}
\Input{$\textnormal{Start point}~\mathbf{x}_{\textnormal{start}}$, \textnormal{goal region}~$X_{\textnormal{goal}}$}
\Output{$\textnormal{Feasible path}~\mathcal{T_F}$}
\textcolor{purple}{$X_{\textnormal{sampled}} \gets \{\xm{start},\mathbf{x}_{\textnormal{goal}}\}$}, $E_\mathcal{F} \gets \emptyset$, $\mathcal{T_F} = (V_\mathcal{F}, E_\mathcal{F})$ \\$\cehism \gets \infty$, $\gamma \gets \infty$, $e_{\gamma} \gets \infty$\;

\While{\textbf{not} $\terminateCondition()$}{
    \textcolor{purple}{$X_{\textnormal{sampled}} \stackrel{+}{\leftarrow} \adaptiveSampler()$}\\
    $\mathcal{T_R}\gets\lazyReverseSearch()$\\
    \While{$\couldImproveForwardSearch(\mathcal{T_R})$}
    {   
        $E_\mathcal{F} \gets\forwardSearch(\mathcal{T_R})$\\

        \eIf{$\isCollide(E_\mathcal{F})$}{
        \textcolor{purple}{$\updateEIS(\cehism, \gamma, e_{\gamma})$}
        % \Comment{Fig.~\ref{fig:generalConcept}}
        \\
    $\updateLazyReverseSearch()$\\

    }{
    
    $\mathcal{T_F}\stackrel{+}{\leftarrow} E_\mathcal{F}$ \Comment{add the valid edges into the tree}\\
        
    }
    
    }$\prune(X_{\textnormal{sampled}})$\\
        }{\Return {$\mathcal{T_F}$}

}
\end{algorithm}
\subsection{Obstacle-Based Adaptive Sampler}\label{subsec: obs_sampler.}
\begin{figure*}[t!] % [h]选项尝试将图片放置在当前位置
    \centering % 让图片居中显示
    \includegraphics[width=0.99\textwidth]{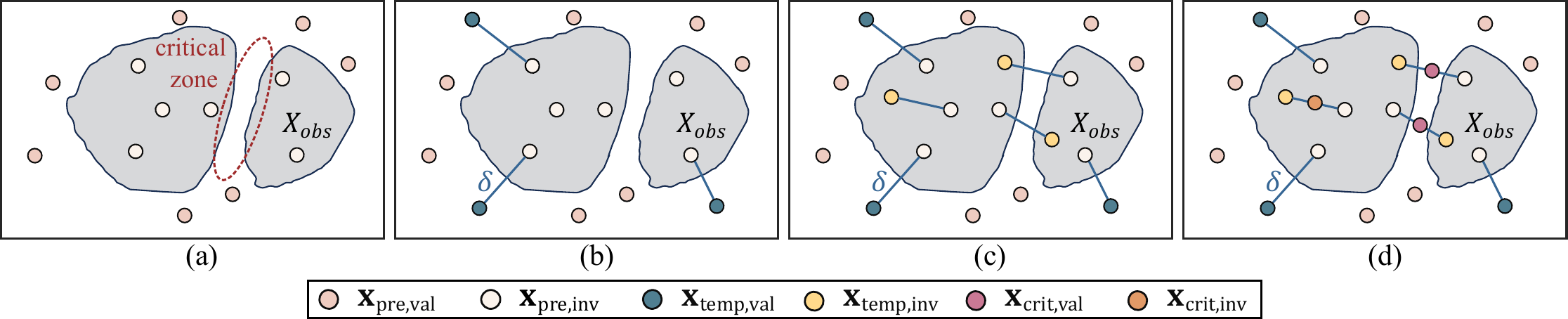} % 调整图片大小并插入图片
    \caption{Four snapshots illustrate how the obstacle-based adaptive sampler adjusts its strategy. (a) depicts uniform sampling in the \textit{$\mathcal{C}$-space} to generate valid/invalid preliminary sampled states (i.e., \x{pre,val} and \x{pre,inv}), with the red dashed ellipsoid highlighting a critical zone which is often difficult to sample. (b) and (c) shows the sampler encountering \x{pre,inv} when using uniform sampling; it employs a Gaussian distribution at distance $\delta$ to find temporary valid/invalid samples (i.e., \x{temp,val} and \x{temp,inv}) around obstacles ($X_{\textnormal{obs}}$). (d) demonstrate that if the \x{temp,inv} fall within $X_{\textnormal{obs}}$, the sampler tests along the connecting bridge between \x{pre,inv} and \x{temp,inv} to sample key points \x{crit,val} in critical zone.}
 % 添加图片标题
    \label{fig:adaptiveSampler} % 为图片添加标签，方便在文中引用
    \vspace{-1.4em} 
\end{figure*}

\begin{algorithm}[t!]
\caption{MIT* - Adaptive sampler \textcolor{black}{(Fig.~\ref{fig:generalConcept})}}
\label{Alg:Sampler}
\DontPrintSemicolon
% \scriptsize
\small
\SetKwInOut{Input}{input}\SetKwInOut{Output}{output}
\SetKwIF{If}{ElseIf}{Else}{if}{}{else if}{else}{end if}%
\SetKwFunction{isValid}{isValid}
\SetKwFunction{sampleSpace}{sampleSpace}
\SetKwFunction{sampleGaussian}{sampleGaussian}
\SetKwFunction{sampleEIS}{sampleEIS}
\SetKwFunction{sampleIS}{sampleIS}

\Input{Maximum iteration $k$, Standard deviation $\delta$}
\Output{
$\Xm{sampled}$ that has denser distribution in critical zones}
% \emph{$\Xm{sampled} = \emptyset$}\\
\emph{$\xm{pre} = \vec{0}, \xm{temp} = \vec{0}, \xm{crit} = \vec{0}$}\\
\For{$i = 1 \rightarrow k$}{
\eIf{$\mn{c}{curr} = \infty $ \textnormal{\Comment{no valid solution found}}}{
\eIf{$\cahism = \infty $ \textnormal{\Comment{no lazy solution found (case 1)}}}{\emph{$\xm{pre} \leftarrow \sampleSpace(\Xm{free})$}}
{\textcolor{purple}{\emph{$\xm{pre} \leftarrow \sampleEIS(\xm{start}, \xm{goal}, \cehism)$}}\\
\Comment{sample estimated informed set (case 2)}}}{
\emph{$\xm{pre} \leftarrow \sampleIS(\xm{start}, \xm{goal}, \mn{c}{curr})$}\\
\Comment{sample informed set (case 4)}}
\eIf{$\isValid(\xm{pre})$}
{
\emph{$\Xm{sampled} \stackrel{+}{\leftarrow} \xm{pre}$}
\Comment{if \x{pre} is valid, add it to $\Xm{sampled}$}\\
}
{
\emph{$\xm{temp} \leftarrow \sampleGaussian(\xm{pre},\delta)$\\}
\eIf{$\isValid(\xm{temp})$}{
\emph{$\Xm{sampled} \stackrel{+}{\leftarrow} \xm{temp}$\\}
}{
\For{$\xm{crit} \in (\xm{pre}, \xm{temp})$}
{
\If{$\isValid(\xm{crit})$}{\emph{$\Xm{sampled} \stackrel{+}{\leftarrow} \xm{crit}$\\}
\textbf{break}\Comment{exit until valid \x{crit} is found}
}}}}
}
\Return{$\Xm{sampled}$}\\
\end{algorithm}
%
%
% Adaptive sampler integrates the sampling methods of uniform random sampling and obstacle-based sampling, leveraging the advantages of both to improve the distribution of samples near obstacles and in constrained areas such as gaps in walls or narrow passageways (Alg.~\ref{Alg:Sampler}).
%
\textcolor{black}{The sampling strategy of the adaptive sampler is categorized into distinct scenarios. During sampling, the uniform sampler (Alg.~\ref{Alg:Sampler}, line 3-10) is employed to generate the preliminary sampled state, denoted as $\mathbf{x}_{\textnormal{pre}}$, and for conducting the initial validity assessment.} The Gaussian distribution-generated temporary state (Alg.~\ref{Alg:Sampler}, line 8) is denoted as $\mathbf{x}_{\textnormal{temp}}$. The state within the critical zone is denoted as $\mathbf{x}_{\textnormal{crit}}$. \textcolor{black}{The adaptive sampling process for valid and invalid states with obstacles-based sampled states distribution is detailed in Fig.~\ref{fig:adaptiveSampler}.}

\begin{itemize}
\item \textit{\x{pre,val}:} Upon validation of preliminary samples \x{pre}, it is incorporated into the resultant sample set $\Xm{sampled}$.
\item \textit{\x{pre,inv}:} For an invalid \x{pre}, a Gaussian distribution is utilized to procure a temporary sampled state \x{temp}, positioned at a $\delta$ distance from \x{pre,inv}.
\item \textit{\x{temp,val}:} If \x{temp} is valid, it is added to set $\Xm{sampled}$ to increase the density of samples close to obstacles.
\item \textit{\x{temp,inv}:} When \x{temp} is invalid, a binary search is conducted to locate a state \x{crit} situated between \x{pre} and \x{temp}, essentially on the edge (\x{pre}, \x{temp}).
\item \textit{\x{crit,val}:} Upon validation of \x{crit}, the first valid \x{crit} is appended to set $\Xm{sampled}$ to enhance the density of states in critical zones (e.g., narrow corridors, wall gaps).
\item \textit{\x{crit,inv}:} If no viable area exists between \x{pre} and \x{temp}, all \x{crit} are invalid, the re-sampling process is recommenced.
\end{itemize}

For the analysis of sample distributions in confined settings. We define the sampling distribution in the critical zone as a weighted mixture of probability functions $\pi_\textnormal{adapt}$. Let $\pdfm{}(\cdot)$ be the probability density of any state $\xm{} \in X_\textnormal{}$, thus:
\begin{equation}
\pdfm{}(\xm{}):=
    \begin{cases}
        1 & \textnormal{if } \xm{} \in \Xm{obs}  \\
        0 & \textnormal{if }  \xm{} \not\in \Xm{obs}
    \end{cases} 
    \hspace{10pt}\textnormal{with}\hspace{10pt}\texttt{Vol}(\Xm{obs}) = 1,
\end{equation}
here, $\texttt{Vol}(\cdot)$ is the \textit{$\mathcal{C}$-space} volume function. \textcolor{black}{The conditional probability density function of \x{temp} given \x{pre} is defined as:
\begin{equation}
\pdfm{}(\xm{temp}|\xm{pre}):=\rho_\textnormal{gau}(\xm{temp})\mathcal{B}(\xm{temp})/\Psi_\textnormal{const},
\label{eq:ftemp}
\end{equation}
where $\rho_\textnormal{gau}(\xm{})$ is the density function of Gaussian distribution around \x{pre}, and $\mathcal{B}(\xm{})$ is a binary function that $\mathcal{B}(\xm{}) = 1$ when $\xm{} \in X_\textnormal{obs}$, and 0 otherwise.
\begin{equation}
\Psi_\textnormal{const} :=  \int_{X}{\rho_\textnormal{gau}(\xm{temp})\mathcal{B}(\xm{temp})}d\xm{temp},
\end{equation}
where $\Psi_\textnormal{const} \in \mathbb{R}$ is a normalizing constant. Thus we have the probability $\pi_\textnormal{adapt}$ created by the adaptive sampler as:
\begin{equation}
\pi_\textnormal{adapt}(\xm{crit}) :=  \int_{X}{\pdfm{}(\xm{temp}|\xm{pre})\pdfm{}(\xm{pre})}d\xm{pre}.
\label{eq:piAdaptive}
\end{equation}
since \x{crit} represents the sample points between line segment $\overline{\xm{pre}\xm{temp}}$, therefore the \x{temp} can be defined as:
\begin{equation}
\xm{temp} := \mn{\xi}{}\cdot\xm{crit} - \xm{pre},   \label{eq:xtemp}
\end{equation}
where $\xi \in \mathbb{R}^{+}$ is defined as the position of the \x{crit}, with its maximum value equal to the resolution of the state space.
By substituting $\pdfm{}(\xm{temp}|\xm{pre})$, $\pdfm{}(\xm{pre})$ and $\xm{temp}$, we have:
\begin{multline}
\pi_\textnormal{adapt}(\xm{crit}) := \\
\int_{X \cap \Xm{obs}}{\frac{\rho_\textnormal{gau}(\mn{\xi}{}\xm{crit} - \xm{pre})\mathcal{B}(\mn{\xi}{}\xm{crit} - \xm{pre})}{\Psi_\textnormal{const}}}d\xm{pre}. 
\end{multline}
where $\pdfm{}(\xm{pre}) = 1$ (preliminary invalid samples). The $\rho_\textnormal{gau}(\xm{})$ is large if  \x{temp} lies close to \x{pre}, and the integrand in Eq.~\ref{eq:piAdaptive} is non-zero only if \x{temp} is invalid. Therefore, for a point \x{crit} in critical zones, the probability density $\pi_\textnormal{adapt}$ is higher. }
\subsection{Estimated Informed Set}\label{subsec: EIS.}
% \begin{figure*}[t!] % [h]选项尝试将图片放置在当前位置
%     \centering % 让图片居中显示
%     \includegraphics[width=0.98\textwidth]{figure/estimatedset/estimatedIS.pdf} % 调整图片大小并插入图片
%     \caption{Diagrammatic depiction of the path planning process of MIT*.} % 添加图片标题
%     \label{fig:EIS} % 为图片添加标签，方便在文中引用
%     \vspace{-1.7em} 
% \end{figure*}
% It first introduces an innovative approach for generating an estimated admissible heuristic solution (\ehis) from an admissible heuristic initial solution (\ahis) that facilitates the shift from a failed forward search to a restarted reverse search. It strategically leverages the lengths of pre-established trees to guide the sampling process toward critical regions of the given scenario, such as narrow passages or gaps.

% \subsubsection{Informed set}
%
\begin{algorithm}[t]
\caption{MIT* - Update estimate informed set}
\label{alg: updateEIS}
\DontPrintSemicolon
\small

\SetKwInOut{Input}{input}\SetKwInOut{Output}{output}
\SetKwIF{If}{ElseIf}{Else}{if}{}{else if}{else}{end if}%
\SetKwFunction{updatedEstimatedSamples}{updatedEstimatedSamples}
\SetKwFunction{computeReliability}{computeReliability}
\SetKwFunction{computeExpandFactor}{computeExpandFactor}
\SetKwFunction{updateEIS}{updateEIS}
\SetKwProg{Function}{Function}{}{}

% \Input{Maximum attempts $n$, Standard deviation $\delta$}
% \Output{Samples $V$ that has denser distribution in critical zones}

% \emph{$\cahism \leftarrow \infty , \cehism \leftarrow \infty$ }\\

\Input{Estimated initial cost \cehis, reliability of admissible initial solution $\gamma$,  expansion
factor $e_{\gamma}$}
\Output{Updated estimated initial cost \cehis}
\Function{$\updateEIS($\cehis$, \gamma, e_{\gamma})$}{
\eIf{\cehis $\neq$ $\infty$}{
\emph{$\cehism \leftarrow \cehism \cdot e_{\gamma}$}\Comment{if the \cehis already exist, but the planner cannot find a valid path within it, adjust the \cehis for estimated informed set expansion (case 3)}\\
}{
\emph{$\cahism \leftarrow g_{\mathcal{F}}(\mathbf{x}_s) + \hat{c}(\mathbf{x}_s,\mathbf{x}_t) + \hat{h}(\mathbf{x}_t)$}\\\Comment{compute \cahis if \cehis does not exist}\\
\emph{$\gamma \leftarrow \computeReliability(\cahism)$ }\Comment{Eq.~\ref{eq:realiability}}\\
\emph{$e_{\gamma} \leftarrow \computeExpandFactor(\gamma)$}\Comment{Eq.~\ref{eq:expandFactor}}\\
%\emph{$\gamma \leftarrow \frac{g_{\mathcal{F}}(\mathbf{x}_s)}{g_{\mathcal{F}}(\mathbf{x}_s) + %\hat{c}(\mathbf{x}_s,\mathbf{x}_t) + \hat{h}(\mathbf{x}_t)} $}\\
%\emph{$e_{\gamma} \leftarrow \sqrt{1+(1-\gamma)}$}\\
\emph{$\cehism \leftarrow \cahism \cdot e_{\gamma}$ \Comment{\textnormal{calculate to define initial \eis}}}\\
}

\Return{\cehis}
}

\end{algorithm}
To expedite the search process, MIT* introduces the concept of the admissible initial solution (\ahis), which is constructed based on the $\mathcal{T_F}$ that have undergone a full collision check and failed (Alg.~\ref{alg: mit}, line 9). In this case, the forward tree is connected with the reverse tree by the edge that failed to pass the full collision check. Although this is a failed search attempt, its cost, denoted as \cahis, can still serve as a valuable heuristic for our path planning problem. A comprehensive definition of the cost of \ahis is provided as follows:
\begin{equation}
\cahism := g_{\mathcal{F}}(\mathbf{x}_s) + \hat{c}(\mathbf{x}_s,\mathbf{x}_t) + \hat{h}(\mathbf{x}_t),
\end{equation}
where $g_{\mathcal{F}}$ is the actual cost to come, $\hat{c}(\mathbf{x}_s,\mathbf{x}_t)$ is the edge cost, and $\hat{h}(\mathbf{x}_t)$ is the admissible cost to reach the goal. Traditional informed set construction delays sampling by requiring the exact solution cost. The \ahis approach enables an estimated informed set to guide sampling before the initial solution is found. However, directly using \ahis can overly restrict the estimated informed set, risking the omission of the exact solution. \textcolor{black}{We define the estimated initial solution \ehis, which is calculated using an expansion factor \(e_{\gamma} > 1\) to expand the sampling range, and a reliability parameter \(\gamma\) to determine \cahis, where:
\begin{equation}
\label{eq:realiability}
\gamma := \frac{g_{\mathcal{F}}(\mathbf{x}_s)}{g_{\mathcal{F}}(\mathbf{x}_s) + \hat{c}(\mathbf{x}_s, \mathbf{x}_t) + \hat{h}(\mathbf{x}_t)} = \frac{g_{\mathcal{F}}(\mathbf{x}_s)}{\cahism}.
\end{equation}
The expansion factor is then computed as:
\begin{equation}
\label{eq:expandFactor}
e_{\gamma} := \sqrt{1 + (1 - \gamma)^2},
\end{equation}
which serves to adjust the extent of the sampling range based on the reliability of the initial estimate. The optimal of $e_{\gamma}$ could be a potential direction for future work.
}
This yields a hypothetical path \ehis, whose cost is denoted as \cehis and can be calculated as follows:
\begin{equation}
{\cehism := \cahism \cdot e_{\gamma},}
\label{eq:cehis}
\end{equation}
with \cehis, we can formulate the estimated informed set with the detailed definition provided in Fig.~\ref{fig:generalConcept} and Alg.~\ref{alg: updateEIS}.
 
An estimated informed set, $\eism$, represents all states within \X{free}, where any state $\mathbf{x} \in \eism$ can potentially be part of a path that has a lower cost than the cost of the current \ehis. This set can be mathematically described by the equation:
\begin{equation}
    \eism := \{ \mathbf{x} \in \Xm{free} \mid \Tilde{f}(\mathbf{x}) < \cehism \},
\end{equation}
where $\Tilde{f}(\mathbf{x}) := \Tilde{g}(\mathbf{x}) + \Tilde{h}(\mathbf{x})$ combines the cost to reach state $\mathbf{x}$ from the start $\Tilde{g}(\mathbf{x})$ and the cost from state $\mathbf{x}$ to the goal $\Tilde{h}(\mathbf{x})$. 
% The benefit of an estimated informed set arises from its ability to efficiently direct sampling efforts towards areas of the state space that are more likely to yield an initial solution.
% 
For problems where the objective is to minimize path length, the estimated informed set typically takes the shape of an $n$-dimensional prolate hyperspheroid or an ellipsoid that is elongated along one axis, which encompasses all points that can result in a path length less than \cehis. The $n$-dimensional hyperellipsoid is defined by the initial position \x{start} and terminal position \x{goal}, alongside two critical metrics: the cost of the \ehis so far, \cehis and conjugate diameters of $\sqrt{\cehism^2 - \sminm^2}$,
where the minimal possible cost \smin comes from:
\begin{equation}
    \sminm := \| \xm{goal} - \xm{start}\|_2,
\end{equation}
here, the eccentricity of the ellipse is given by ratio $\sminm/\cehism$. 
The estimated informed set is the intersection of the free space,
\X{free}, and an $n$-dimensional Hyper-Ellipsoid (\X{HES}) which symmetric about its transverse axis:
\begin{equation}
    \eism = \Xm{free} \cap \Xm{HES},
\end{equation}
where
\begin{equation}
\Xm{HES} := \{\xm{} \in \mathbb{R}^n \mid \|\xm{} - \xm{start}\|_2 + \|\xm{goal} - \xm{}\|_2 < \cehism
\} .
\end{equation}

The estimated informed set \eis in $\mathbb{R}^2$ can be directly defined using the $L^2$ norm (i.e., Euclidean distance):
\begin{equation}
    \eism := \{ \xm{} \in \Xm{free} \mid \|\xm{} - \xm{start}\|_2 + \|\xm{goal} - \xm{}\|_2 < \cehism \},
\end{equation}
\begin{algorithm}[t!]
\caption{MIT* - Sample estimated informed set}
\label{alg:sampleEIS}
\DontPrintSemicolon
% \scriptsize
\small
\SetKwInOut{Input}{input}\SetKwInOut{Output}{output}
\SetKwIF{If}{ElseIf}{Else}{if}{}{else if}{else}{end if}%
\SetKwFunction{sampleUnitBall}{sampleUnitBall}
\SetKwFunction{SVD}{SVD}
\SetKwFunction{sampleEIS}{sampleEIS}
\SetKwFunction{getDim}{getDim}
\SetKwProg{Function}{Function}{}{}

\Input{Start $\xm{start}$, goal $\xm{goal}$, estimate initial cost $\cehism$}
\Output{$\Xm{sampled}$ within estimated informed set}

\Function{$\sampleEIS(\xm{start}, \xm{goal}, \cehism)$}{

\emph{$\Xm{sampled} = \emptyset$}\\
\emph{$n = \getDim(\xm{start})$}\\
\emph{$\sminm \leftarrow \|\xm{} -\xm{start}\|_2$}\\
\emph{$\xm{center} \leftarrow (\xm{start} + \xm{goal})/2$}\\
\emph{$\mb{k}{1} \leftarrow (\xm{goal} - \xm{start})/\sminm$}\Comment{major axis direction}\\
\emph{$\{\mb{U}{},\mb{V}{}\} \leftarrow \SVD(\mb{k}{1}\mb{I}{1}^T)$}\Comment{unitary singular matrices}\\
\emph{$\mb{\Lambda}{} \leftarrow diag(1,...,1,det(\mb{U}{})det(\mb{V}{}))$}\\
% \Comment{Ensures R has positive determinant}\\
\emph{$\mb{R}{} \leftarrow \mb{U}{}\mb{\Lambda}{}\mb{V}{}^T$}\Comment{rotation matrix via decomposition}\\
\emph{$\mn{l}{1} \leftarrow \cehism/2$}\Comment{use \cehis as the length of major axis}\\
\emph{$\{\mn{l}{j}\}_{j=2,...,n} \leftarrow (\sqrt{\cehism^2-\sminm^2})/2$}\Comment{set other axis' length}\\
\emph{$\mb{L}{} \leftarrow diag(\mn{l}{1},\mn{l}{2},...,\mn{l}{n})$}\\
\For{$i = 1 \rightarrow k$}{
\emph{$\xm{ball} \leftarrow \sampleUnitBall(n)$\\}
\emph{$\xm{hes} \leftarrow \mathbf{R}\mb{L}{}\xm{ball} +\xm{center}$\\}
\emph{$\Xm{sampled} \stackrel{+}{\leftarrow} \xm{hes}$\\}}
\Return{$\Xm{sampled}$}\\
}
\end{algorithm}

\textcolor{black}{To extend EIS into multiple goals scenarios based on informed set~\cite{gammell2018informed}, goal region $X_\textnormal{goal}:=\left\{\xm{goal,o}\right\}_{o=1}^m$ set is the union of the individual estimated informed sets:
\begin{equation}
\eism=\bigcup_{o=1}^mX_{\Tilde{\sigma},o},
\end{equation}
where $m$ is the number of goals in the problem domain, therefore the $o$-th EIS $X_{\Tilde{\sigma},o}$ can be described as:
\begin{equation}
    X_{\Tilde{\sigma},o}:= \{ \xm{} \in \Xm{free} \mid \|\xm{} - \xm{start}\|_2 + \|\xm{goal,o} - \xm{}\|_2 < \Tilde{s}_\textnormal{est,o} \},
\end{equation}}
The direct method to generate distributed samples via adaptive sampler in the estimated informed set is defined from direct sampling in subset~\cite{gammell2014informed}.
We define the $n$-dimensional hyper ellipsoid that contains an estimated informed set as:
\begin{equation}
    \Xm{HES} := \{ \xm{} \in \mathbb{R}^n \mid (\xm{} -\xm{center})^T\mb{R}{}\mb{P}{}^{-1}\mb{R}{}^T(\xm{} -\xm{center}) < 1\},
\end{equation}
where
\begin{equation}
    \xm{center} := \frac{\xm{start}+\xm{goal}}{2},
\end{equation}
\begin{equation}
    \mb{P}{} := diag\left(\frac{\cehism^2}{4},\frac{\cehism^2-\sminm^2}{4},...,\frac{\cehism^2-\sminm^2}{4}\right),
\end{equation}
where $\mb{P}{} \in \mathbb{R}^{n\cross n}$ is the positive symmetric matrix, and the rotation matrix \tb{R}{} can be obtained by \textit{minimizing a cost function} that measures the alignment error between the rotated hyperellipsoid's major axis and the first axis of the world frame. The cost function is typically defined as:
\begin{equation}
    \mn{J}{}(\mb{R}{}) := \|\mb{R}{}\mb{k}{1}-\mb{I}{1}\|^2,
\end{equation}
where $\mb{k}{1} \in \mathbb{R}^{n \cross 1}$ (Alg.~\ref{alg:sampleEIS}, line 6) is the major axis of the rotated hyperellipsoid and $\mb{I}{1}\in \mathbb{R}^{n \cross 1}$ is the first column of the identity matrix, which represents the first axis of the world frame. Thus we can retrieve rotation matrix \tb{R}{}:
\begin{equation}
    \mb{R}{} := \mb{U}{}\mb{\Lambda}{}\mb{V}{}^T,
\end{equation}
where diagonal matrix $\mb{\Lambda}{} \in \mathbb{R}^{n\cross n}$ is:
\begin{equation}
    \mb{\Lambda}{}:=
    \begin{cases}
    \mb{I}{} & \textnormal{if } det(\mb{U}{}\mb{V}{}^T) = 1, \\
    diag(1,1,...,-1) & \textnormal{if }  det(\mb{U}{}\mb{V}{}^T) = -1.
    \end{cases},
\end{equation}
where $det(\cdot)$ is the matrix determinant. The unitary matrices $\mb{U}{} \in \mathbb{R}^{n\cross n}$ and $\mb{V}{} \in \mathbb{R}^{n\cross n}$ are defined by the singular value decomposition $\texttt{SVD}(\cdot)$, such as $\mb{U}{}\mb{\Sigma}{}\mb{V}{}^T \equiv \mb{B}{}$, and $\mb{B}{} \in \mathbb{R}^{n\cross n}$ is calculated by the outer product of axes:
\begin{equation}
    \label{eq:MatrixB}
    \mb{B}{} := \mb{k}{1}\mb{I}{1}^T = ((\xm{goal} - \xm{start})/\sminm)\mb{I}{1}^T.
\end{equation}
with predefined matrices, we can generate samples directly by transforming samples $\xm{ball} \in \Xm{ball}$ into a $\xm{hes} \in \Xm{HES}$. The \tn{X}{ball} is defined as:
\begin{equation}
    \Xm{ball} := \{ \xm{} \in \mathbb{R}^n \mid \|\xm{} \|_2 < 1\},
\end{equation}
and the transform is given by:
\begin{equation}
    \xm{hes} = \mb{R}{}\mb{L}{}\xm{ball} + \xm{center},
\end{equation}
where $\mb{L}{} \in \mathbb{R}^{n\cross n}$ is the lower-triangular Cholesky decomposition of the \tb{P}{} such that: 
\begin{equation}
    \mb{L}{}\mb{L}{}^T\equiv\mb{P}{},
\end{equation}
the \tb{L}{} can be written as:
\begin{equation}
    \mb{L}{} := diag\left(\frac{\cehism}{2},\frac{\sqrt{\cehism^2-\sminm^2}}{2},...,\frac{\sqrt{\cehism^2-\sminm^2}}{2}\right).
\end{equation}

This section defines the \eis which re-projects \textit{$\mathcal{C}$-space} set to the estimated hyperellipsoid set, and the pseudocode of direct sampling within the estimated informed set is given in Alg.~\ref{alg:sampleEIS}.

\subsection{Length-related Adaptive Sparse Collision Check}\label{subsec: approx.}
Traditional sparse collision-checking methods are inefficient due to the fixed number of checks, $\Theta_\textnormal{sparse,$k$} \in \mathbb{N}^+$, to all edges. These methods eventually adapt, but their initial check level is not robust (i.e., only one check per edge), leading to inaccuracies. In contrast, MIT* introduces a \textit{length-related adaptive sparse collision-checking} strategy. MIT* starts with an initial resolution, $\Delta_{\textnormal{sparse,ini}}$ (e.g., $5 \times 10^{-6}$ in simulation benchmark), and adjusts it according to the edge length, adapting the resolution for sparse collision checks as follows:
\textcolor{black}{\begin{align}
    \Delta_{\textnormal{sparse,$k$}} &:= \Omega \cdot \Delta_{\textnormal{sparse,ini}}, \\
    \Theta_\textnormal{sparse,$k$} &:= \lfloor \frac{\| \xm{s} - \xm{t}\|_2}{\Delta_{\textnormal{sparse,$k$}}} + 1\rfloor.
\end{align}
here, $\lfloor \cdot \rfloor$ is the floor function for integers, $\Delta_{\textnormal{sparse,$k$}}$ is the next sparse-check resolution, $\Omega \in (0, 1)$ is the tuning parameter. }
% The edge check number 
The number of edge collision checks $\Theta_\textnormal{sparse,k}$ in $\mathcal{T_R}$ increases when the edge length is longer. This approach enables MIT* to perform effective initial collision checks and adjust resolution, making the reverse search more reliable and problem-specific.

\subsection{Probabilistic Completeness and Asymptotic Optimality}

Most sampling-based path planning algorithms have been proven to be probabilistically complete and asymptotically optimal, and MIT* can also guarantee these two properties. As the number of iterations $k$ approaches infinity, the entire space will be explored, satisfying the following condition:
\begin{equation}
\lim_{k \to \infty} \mathbb{P} (\{V_\mathcal{F}\cup V_\mathcal{R}\} \cap X_{\textnormal{goal}}) \neq \emptyset) = 1, 
\end{equation}
which means that if there is a feasible path, it must be found by the MIT*. Therefore, the probabilistic completeness of MIT* is guaranteed. 

The MIT* implements the same Choose Parent and Rewire strategies as the EIT*. It means that if the rewiring radius $r(q)$  in Choose Parent and Rewire processes satisfies:
\begin{equation}
\label{eqn:radius r}
    r(q) > \eta \left(2 \left(1 + \frac{1}{n}\right){\left(\frac{\lambda(\eism \cup X_{\hat{f}})}{\zeta_n}\right) \left( \frac{\log(q)}{q}\right)}\right)^{\frac{1}{n}},
\end{equation}
here, $q$ denotes the number of sampled states in the sets, $\eta > 1$ is a tuning parameter, $n$ is the dimensionality of the workspace. The term $\lambda(\eism \cup X_{\hat{f}})$ represents the Lebesgue measure of the union of the estimated informed set \eis, and the informed set $X_{\hat{f}}$, where \eis is used in \textit{initial pathfinding} phase and $X_{\hat{f}}$ is used for \textit{path optimization} phase. $\zeta_n$ is the volume of the unit ball in the current workspace. In reference to Lemma 56, 71, and 72 in~\cite{karaman2011sampling}, the following equation holds:
\begin{equation}
\mathbb{P} (\limsup_{q \to \infty} \min_{\sigma\in\Sigma_q} \left\{ c(\sigma) \right\} = c^*) = 1.
\end{equation}
where $q$ is the number of samples, $\Sigma_q \subset \Sigma$ is the set of valid paths from the start to the goal found by the planner from those samples, $c: \Sigma \rightarrow [0, \infty)$ is the cost function, and $c^*$ is the optimal solution cost. It indicates that the MIT* can find an optimal path, if it exists, as the number of iterations go to infinity. Therefore, the asymptotic optimality is guaranteed.
\section{Experimental Results}\label{sec:Expri}
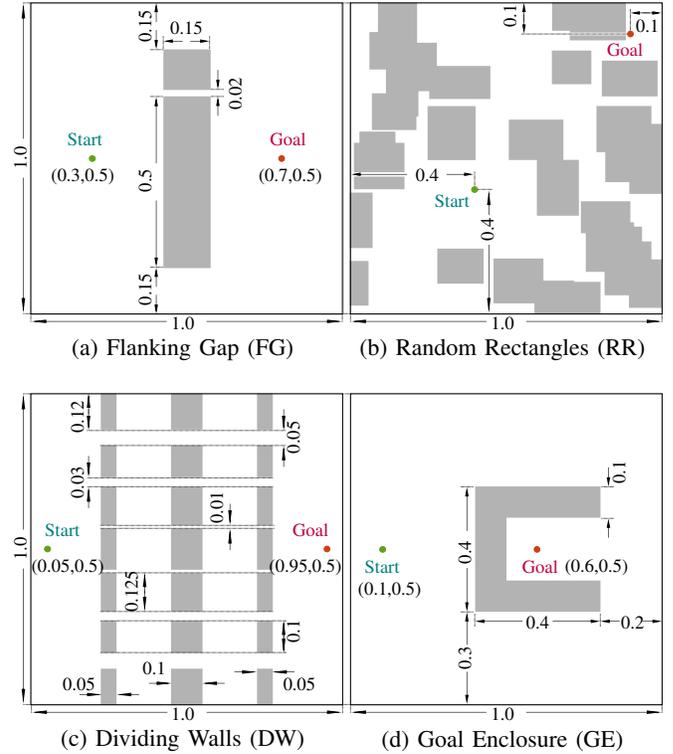
\begin{figure}[t!]
    \centering
    \begin{tikzpicture}
    \node[inner sep=0pt] (russell) at (-4.0,0.0)
    {\includegraphics[width=0.24\textwidth]{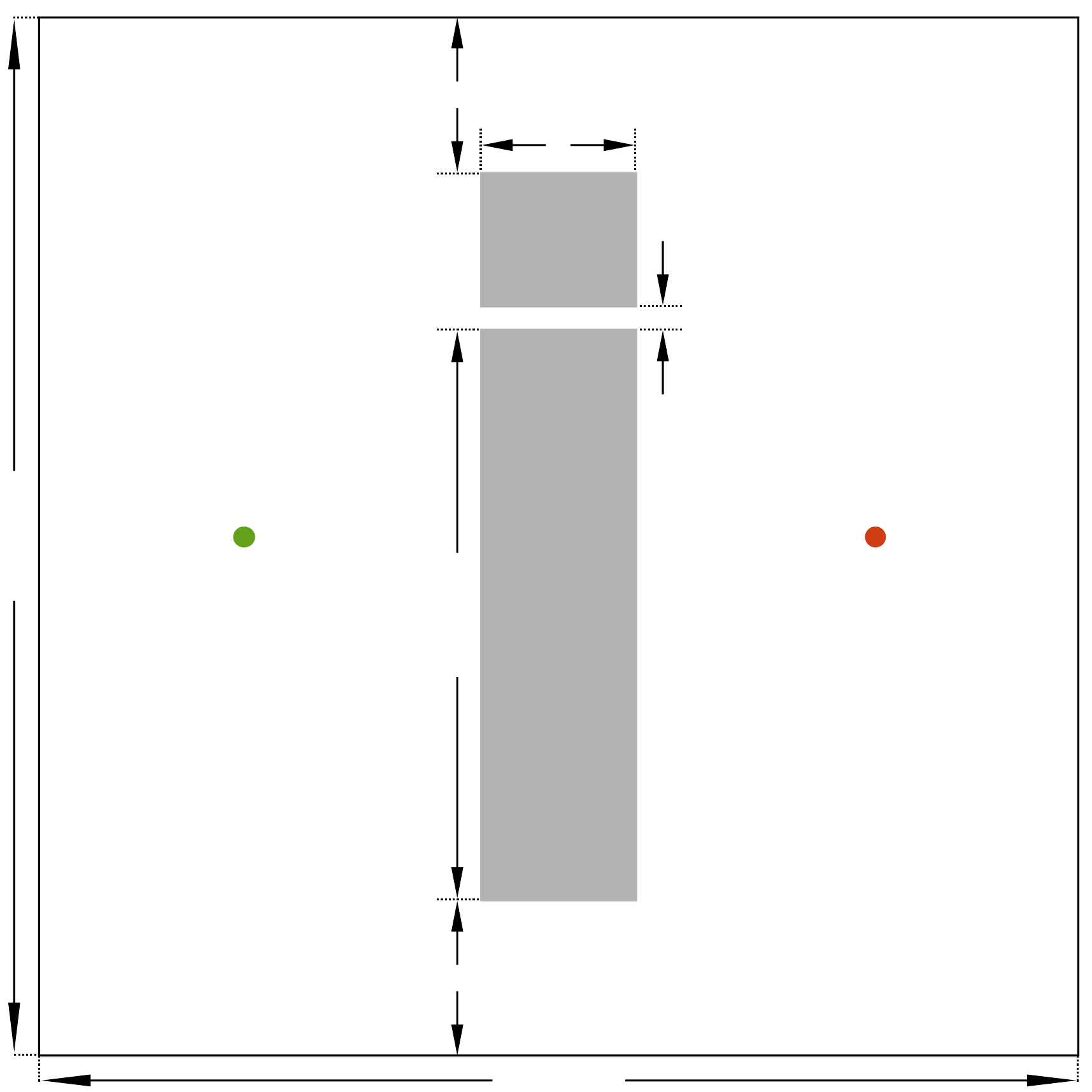}};
    \node[inner sep=0pt] (russell) at (0.25,0.0)
    {\includegraphics[width=0.24\textwidth]{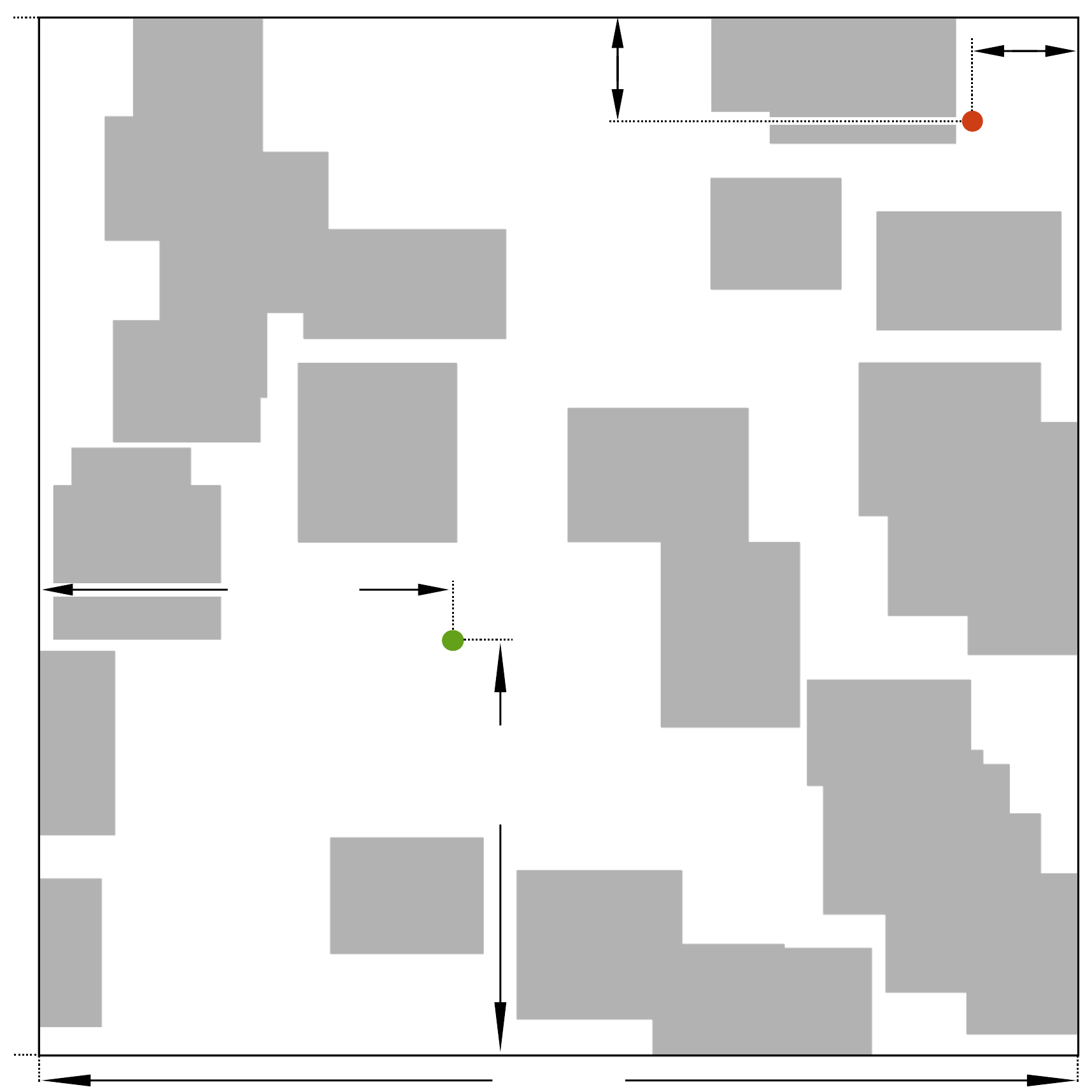}};
    \node[inner sep=0pt] (russell) at (-4.0,-5.2)
    {\includegraphics[width=0.24\textwidth]{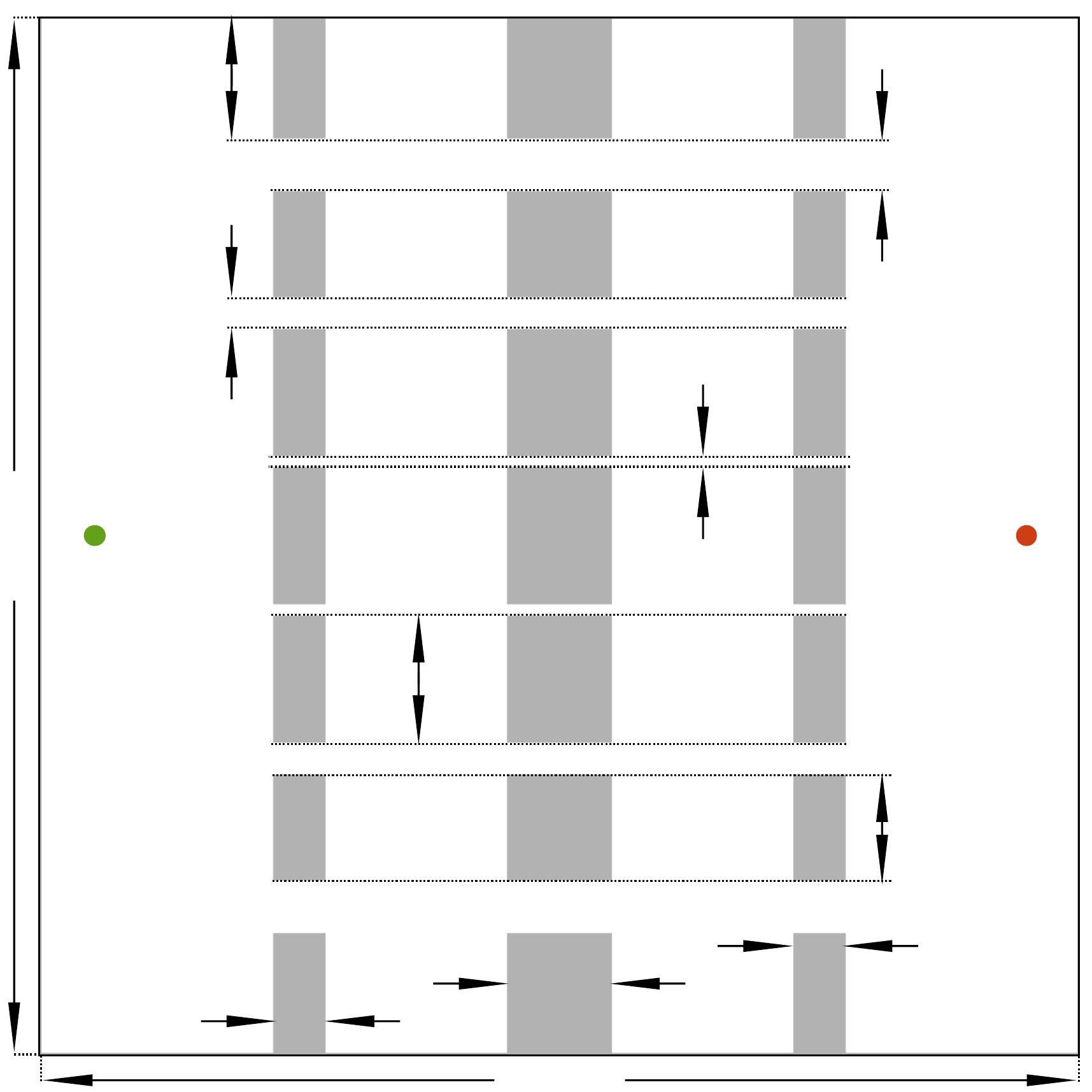}};
    \node[inner sep=0pt] (russell) at (0.25,-5.2)
    {\includegraphics[width=0.24\textwidth]{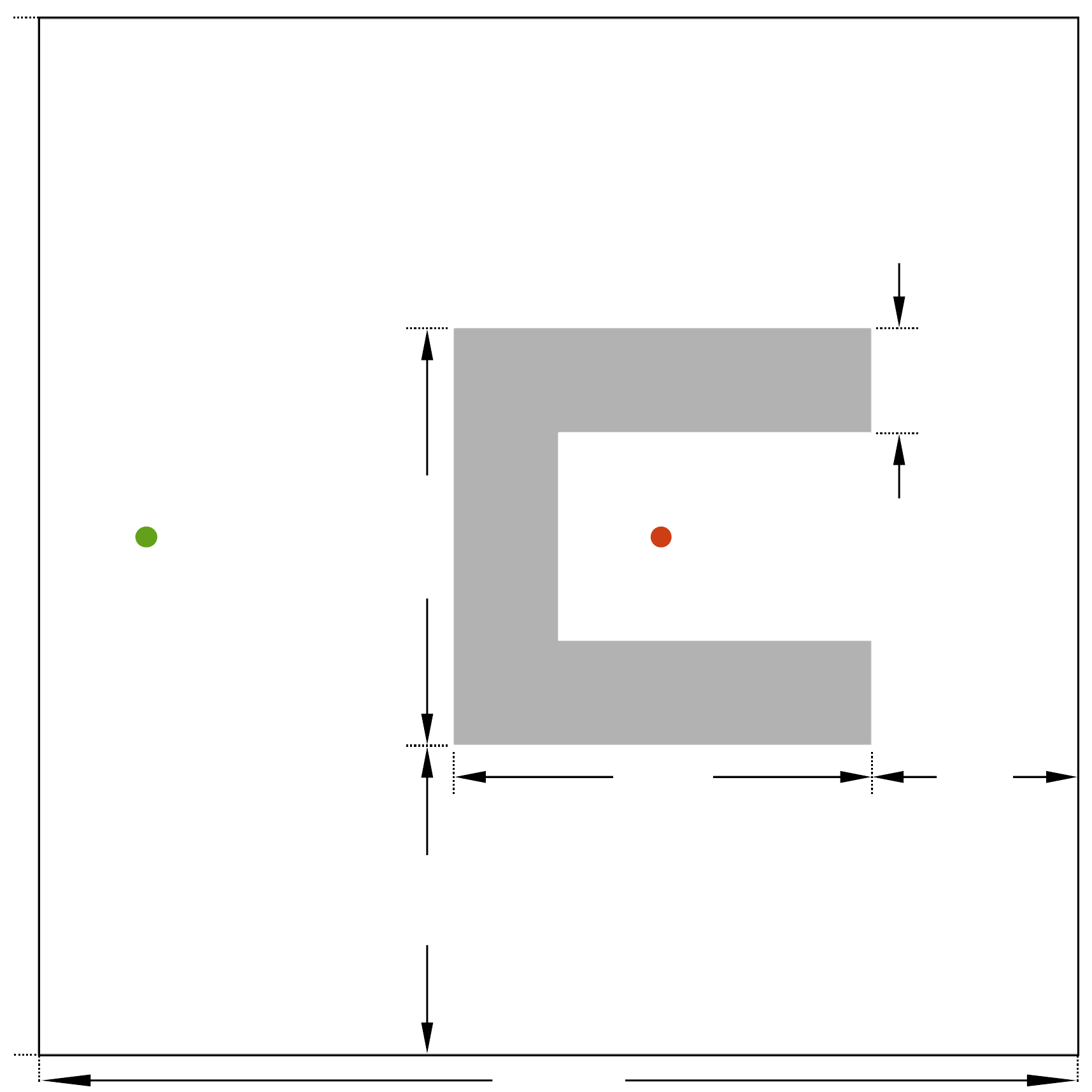}};
    \scriptsize

    %%flanking Gap
    %%left
 
    \node [rotate=90] at (-4.5,-0.25) {0.5};    
    \node [rotate=90] at (-4.5,1.8) {0.15};
    \node at (-3.95,1.78) {0.15};
    % \node at (-5.62,-0.46) {0.05};
    %%right

    \node [rotate=90] at (-3.3,0.9) {0.02};

    % \node [rotate=90] at (-1.92,-1.01) {0.5};
    % \node at (-2.12,0.55) {0.05};
    \node [rotate=90] at (-4.5,-1.7) {0.15};

    %%frame
    \node [rotate=90] at (-6.16,0.05) {1.0};
    \node at (-4.0,-2.15) {1.0};
    \node at (-5.33,-0.2) {(0.3,0.5)};
    \node at (-2.57,-0.2) {(0.7,0.5)};
    \node at (-5.3,0.3) {\color{teal} Start};
    \node at (-2.6,0.3) {\color{purple} Goal};

    %% Random Rectangles
    \node at (-0.75,-0.18) {0.4};
    \node at (0.25,-2.15) {1.0};
    \node [rotate=90] at (0.05,-0.9) {0.4};
    \node [rotate=90] at (0.35,1.94) {0.1};
    \node  at (2.18,1.8) {0.1};

    \node at (-0.42,-0.52) {\color{teal} Start};
    \node at (1.88,1.48) {\color{purple} Goal};
    
    % %% dividing walls
    \node [rotate=90] at (-5.4,-4.27) {0.03};    
    \node [rotate=90] at (-4.68,-5.72) {0.125};    
    \node [rotate=90] at (-5.4,-3.4) {0.12};
    
    \node [rotate=90] at (-2.53,-6.25) {0.1};
    \node [rotate=90] at (-3.55,-4.6) {0.01};
    \node [rotate=90] at (-2.53,-3.67) {0.05};
    
    \node at (-5.38,-6.95) {0.05};
    \node at (-4.38,-6.75) {0.1};
    \node at (-2.45,-6.95) {0.05};
    
    \node [rotate=90] at (-6.16,-5.15) {1.0};
    \node at (-4.0,-7.35) {1.0};
    \node at (-5.53,-5.4) {(0.05,0.5)};
    \node at (-2.37,-5.4) {(0.95,0.5)};
    
    \node at (-5.6,-4.9) {\color{teal} Start};
    \node at (-2.3,-4.9) {\color{purple} Goal};
    
    % %% goal enclosure
    \node at (0.72,-6.13) {0.4};
    \node at (0.25,-7.35) {1.0};
    % \node [rotate=90] at (-1.91,-5.15) {1.0};
    \node [rotate=90] at (1.8,-4.13) {0.1};
    \node [rotate=90] at (-0.25,-5.2) {0.4};
    \node [rotate=90] at (-0.25,-6.61) {0.3};
    
    \node at (1.98,-6.13) {0.2};
    \node at (-1.25,-5.7) {(0.1,0.5)};
    \node at (1.5,-5.4) {(0.6,0.5)};
    
    \node at (-1.35,-5.4) {\color{teal} Start};
    \node at (0.75,-5.4) {\color{purple} Goal};

    %%%%%
    \node at (-4.0,-2.51) {\small (a) Flanking Gap (FG)};
    \node at (0.25,-2.51) {\small (b) Random Rectangles (RR)};
    \node at (-4.0,-7.71) {\small (c) Dividing Walls (DW)};
    \node at (0.25,-7.71) {\small \textcolor{black}{(d) Goal Enclosure (GE)}};
    \end{tikzpicture}
    % \vspace{-0.5em} 
    \caption{The 2D representation of the simulated planning problems in Section~\ref{sec:Expri}. The state space, denoted as $X \subset \mathbb{R}^n$, is constrained within a hypercube with one width for both problem instances. Specifically, we conducted ten distinct instantiations of the random rectangles experiment and the outcomes are showcased in Fig.~\ref{fig: result}.}
    \label{fig: testEnv}
    \vspace{-1.7em} 
\end{figure}
\begin{figure*}[t!]
    \centering
    \begin{tikzpicture}
    %first
    \node[inner sep=0pt] (russell) at (-5.7,8)
    {\includegraphics[width=0.31\textwidth]{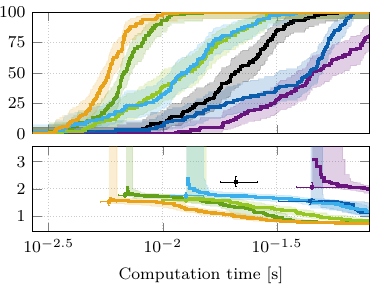}};
    \node[inner sep=0pt] (russell) at (0,8)
    {\includegraphics[width=0.31\textwidth]{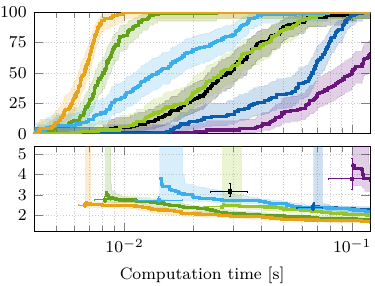}};  
    \node[inner sep=0pt] (russell) at (5.7,8)
    {\includegraphics[width=0.31\textwidth]{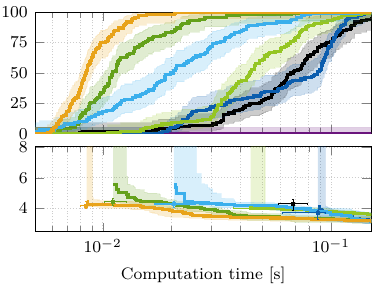}};  

    %second
    \node[inner sep=0pt] (russell) at (-5.7,3.3)
    {\includegraphics[width=0.31\textwidth]{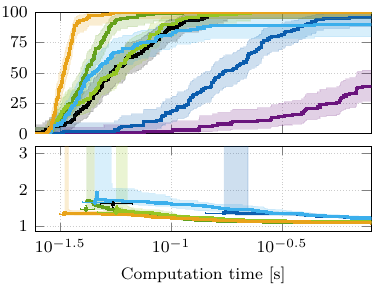}};
    \node[inner sep=0pt] (russell) at (0,3.3)
    {\includegraphics[width=0.31\textwidth]{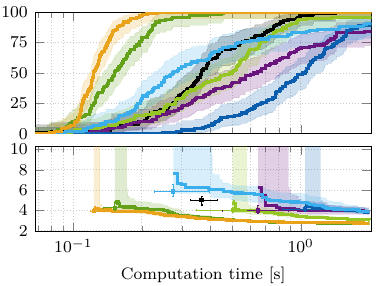}};  
    \node[inner sep=0pt] (russell) at (5.7,3.3)
    {\includegraphics[width=0.31\textwidth]{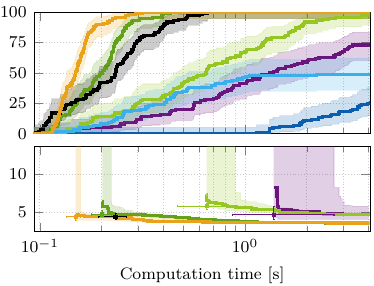}};

    %third
    \node[inner sep=0pt] (russell) at (-5.7,-1.4)
    {\includegraphics[width=0.31\textwidth]{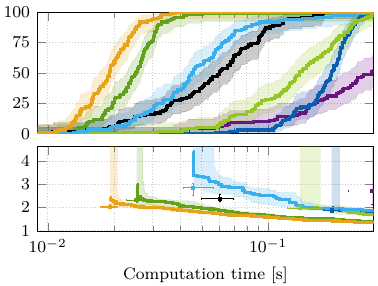}};
    \node[inner sep=0pt] (russell) at (0,-1.4)
    {\includegraphics[width=0.31\textwidth]{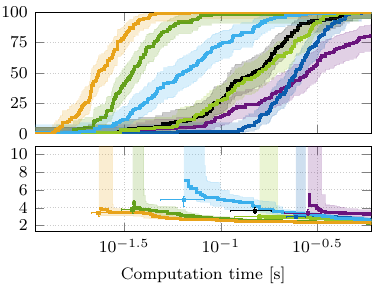}};  
    \node[inner sep=0pt] (russell) at (5.7,-1.4)
    {\includegraphics[width=0.31\textwidth]{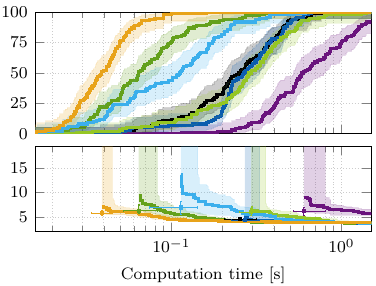}}; 

    %forth
    \node[inner sep=0pt] (russell) at (-5.7,-6.1)
    {\includegraphics[width=0.31\textwidth]{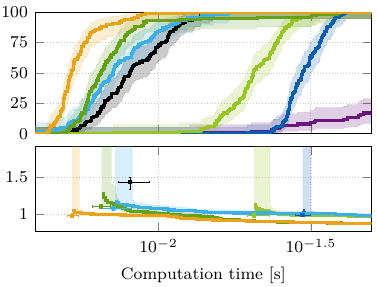}};
    \node[inner sep=0pt] (russell) at (0.1,-6.1)
    {\includegraphics[width=0.32\textwidth]{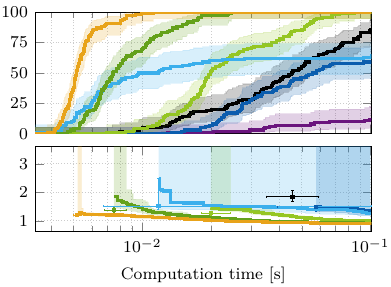}};  
    \node[inner sep=0pt] (russell) at (5.7,-6.1)
    {\includegraphics[width=0.31\textwidth]{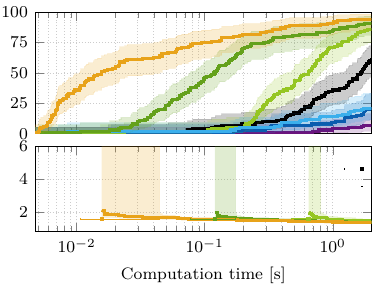}};

    \node[inner sep=0pt] (russell) at (-8.8,8.4){\includegraphics[width=0.025\textwidth]{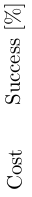}};
    \node[inner sep=0pt] (russell) at (-8.8,3.6){\includegraphics[width=0.025\textwidth]{figure/benchmark/secc_cost.pdf}};
    \node[inner sep=0pt] (russell) at (-8.8,-1.1){\includegraphics[width=0.025\textwidth]{figure/benchmark/secc_cost.pdf}};
    \node[inner sep=0pt] (russell) at (-8.8,-5.8){\includegraphics[width=0.025\textwidth]{figure/benchmark/secc_cost.pdf}};
    
    \node[inner sep=0pt] (russell) at (0.0,-9.2){\includegraphics[width=0.8\textwidth]{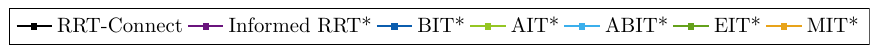}};
    
    %%first
    \node at (-5.5,5.7) {\footnotesize (a) FG in $\mathbb{R}^4$ - MaxTime: 0.08s};
    \node at (0.2,5.7) {\footnotesize (b) FG in $\mathbb{R}^8$ - MaxTime: 0.12s};
    \node at (5.9,5.7) {\footnotesize (c) FG in $\mathbb{R}^{16}$ - MaxTime: 0.20s};

    %%second
    \node at (-5.5,1.0) {\footnotesize (d) RR in $\mathbb{R}^4$ - MaxTime: 0.80s};
    \node at (0.2,1.0) {\footnotesize (e) RR in $\mathbb{R}^8$ - MaxTime: 2.00s};
    \node at (5.9,1.0) {\footnotesize (f) RR in $\mathbb{R}^{16}$ - MaxTime: 4.00s};

    %%third
    \node at (-5.5,-3.7) {\footnotesize (g) DW in $\mathbb{R}^4$ - MaxTime: 0.30s};
    \node at (0.2,-3.7) {\footnotesize (h) DW in $\mathbb{R}^8$ - MaxTime: 0.60s};
    \node at (5.9,-3.7) {\footnotesize (i) DW in $\mathbb{R}^{16}$ - MaxTime: 1.50s};

    %%forth
    \textcolor{black}{
    \node at (-5.5,-8.4) {\footnotesize (j) GE in $\mathbb{R}^2$ - MaxTime: 0.05s};
    \node at (0.2,-8.4) {\footnotesize (k) GE in $\mathbb{R}^4$ - MaxTime: 0.10s};
    \node at (5.9,-8.4) {\footnotesize (l) GE in $\mathbb{R}^{8}$ - MaxTime: 2.00s};}
    
    \end{tikzpicture}
    \vspace{-0.4em} 
    \caption{Detailed experimental results from Section~\ref{subsec:experi} are summarized. Fig. (a)-(c) show flanking gap outcomes in $\mathbb{R}^4$, $\mathbb{R}^8$, and $\mathbb{R}^{16}$, respectively. Panel (d)-(f) displays ten random rectangle experiments in $\mathbb{R}^4$, $\mathbb{R}^8$ and $\mathbb{R}^{16}$. Fig. (g)-(i) present dividing walls outcomes in $\mathbb{R}^4$, $\mathbb{R}^8$, and $\mathbb{R}^{16}$. Fig. (j)-(l) present non-convex simulation for goal enclosure test outcomes in $\mathbb{R}^2$, $\mathbb{R}^4$, and $\mathbb{R}^{8}$. In the cost plots, boxes represent solution cost and time, with lines indicating cost progression for an almost surely optimal planner (unsuccessful runs have infinite cost). Error bars provide nonparametric 99\% confidence intervals for solution cost and time.}
    \label{fig: result}
    \vspace{-1.8em}
\end{figure*}
\begin{figure*}[t!]
    \centering
    \begin{tikzpicture}
    \node[inner sep=0pt] (russell) at (0,0)
    {\includegraphics[width=0.94\textwidth]{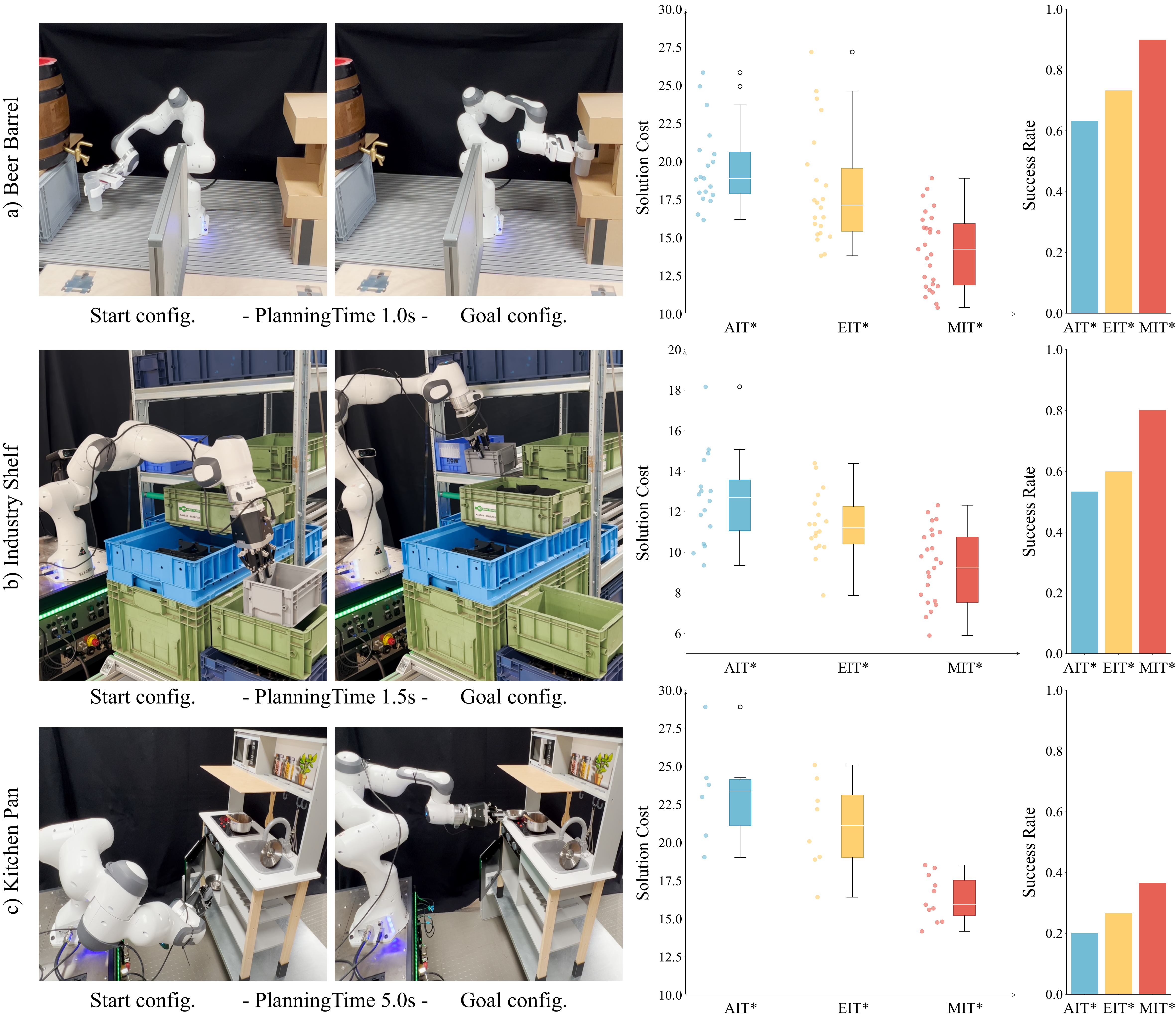}};

    \end{tikzpicture}
    \vspace{-0.3em} 
    \caption{Detailed experimental results from Section~\ref{subsec:realExpri} are summarized above. Fig.~\ref{fig: Realresult}a illustrates the \textit{beer barrel}, highlighting the start and goal configurations, solution cost, and success rate. Fig.~\ref{fig: Realresult}b depicts the \textit{industry shelf}, showing the initial and final positions for container extraction and placement. Fig.~\ref{fig: Realresult}c presents the \textit{kitchen}, focusing on the DARKO robot's performance. In the cost box plots, boxes represent solution costs per planner, while lines show mean cost progression for an optimal planner (with unsuccessful runs assigned an infinite cost).}
    \label{fig: Realresult}
    \vspace{-1.7em}
\end{figure*}
In this article, we utilize the Planner-Arena benchmark database~\cite{moll2015benchmarking}, the Planner Developer Tools (PDT)~\cite{gammell2022planner}, and MoveIt~\cite{gorner2019moveit} to benchmark proposed motion planner behaviors. \textcolor{black}{MIT* was tested against SOTA algorithms in both simulated random scenarios (Fig.~\ref{fig: testEnv}, resolution $5 \times 10^{-6}$) and real-world manipulation problems (Fig.~\ref{fig: Realresult}, resolution $5 \times 10^{-3}$).} The comparison involved several versions of RRT-Connect, Informed RRT*, BIT*, ABIT*, AIT*, and EIT* sourced from the Open Motion Planning Library (OMPL)~\cite{sucan2012open}. \textcolor{black}{The construction time costs of the \textit{$\mathcal{C}$-space} (Table~\ref{tab:benchmark}) across different dimensions are evaluated using the space information constructor of OMPL.
The experimental results demonstrate that the \textit{$\mathcal{C}$-space} construction time remains a minor component of the overall planning process and does not constitute a computational bottleneck.} The evaluations are implemented on a computer with an Intel i7 3.90 GHz processor and 32GB of LPDDR3 3200 MHz memory. These comparisons were carried out in simulated environments ranging from $\mathbb{R}^2$ to $\mathbb{R}^{16}$. The primary objective for the planners was to minimize path length (cost). The RGG constant $\eta$ was uniformly set to 1.001, and the rewire factor was set to 1.2 for all planners. \textcolor{black}{Gaussian standard deviation $\delta$ is default set to 10\% of the maximum extent of \textit{$\mathcal{C}$-space}.}

For RRT-based algorithms, a 5\% goal bias was used, with maximum edge lengths of 0.3, 0.5, 1.25, and 3.0 in $\mathbb{R}^2$, $\mathbb{R}^4$, $\mathbb{R}^8$, $\mathbb{R}^{16}$. All batch-sorted planners sampled 100 states per batch, and informed planners defined the informed set $X_{\hat{f}}$ using the current best costs.
MIT* leveraged prior admissible costs to guide the search using an \textit{estimated informed set} $\eism$ before discovering the initial solution. It then used an adaptive sampler to target critical areas in confined spaces. \textcolor{black}{The implementation of MIT* planner into OMPL framework is available at: }\href{https://github.com/Liding-Zhang/ompl_release}{\textcolor{black}{https://github.com/Liding-Zhang/ompl\_release}}.
\subsection{Simulation Experimental Tasks}\label{subsec:experi}
The planners were tested across four distinct benchmarks in four domains: $\mathbb{R}^2$, $\mathbb{R}^4$, $\mathbb{R}^8$, and $\mathbb{R}^{16}$. \textcolor{black}{In the first scenario, a constrained flanking gap (FG) environment with a narrow gap was simulated, offering one general direction for optimal path (Fig.~\ref{fig: testEnv}a).} Each planner ran 100 times with varying random seeds, and the computation times for optimal planners are labeled. Figs.~\ref{fig: result}a,~\ref{fig: result}b and \ref{fig: result}c show that MIT* quickly finds the initial solution in different dimensions with minimal time.

\textcolor{black}{In the second scenario, random widths were assigned to \textit{axis-aligned hyperrectangles}, creating non-convex regions within the \textit{$\mathcal{C}$-space} (Fig.~\ref{fig: testEnv}b).
Ten random rectangle (RR) problems were generated for each \textit{$\mathcal{C}$-space} dimension, with 100 trials per planner. Figs.~\ref{fig: result}d,~\ref{fig: result}e and \ref{fig: result}f show that MIT* achieved the highest success rates within a limited time}.

\textcolor{black}{The third test simulated a constrained dividing wall (DW) environment with multiple narrow gaps, allowing all planners to search optimal paths in various non-trivial directions (Fig.~\ref{fig: testEnv}c).} Each planner ran 100 times with different random seeds, and maximal computation times for optimal planners are shown in the labels. Figs.~\ref{fig: result}g,~\ref{fig: result}h and \ref{fig: result}i indicate that MIT* outperforms the SOTA planner in both finding the initial solution and converging to the optimal solution.

\textcolor{black}{The last test problem featured a hollow, non-convex C-shaped obstacle surrounding the goal state. In this configuration, even in higher dimensions, the goal could only be accessed through the face of the hyper-obstacle farthest from the start state (Fig.~\ref{fig: testEnv}d). This setup presents a challenge for sampling-based planners, as many invalid edges near the reverse search tree often require repair. Effective coverage of narrow channels is crucial, and strategies such as biased or adaptive sampling may be necessary to improve efficiency. A direct path planner would result in collisions, resulting in the planner rerouting around obstacles. As shown in Figs.~\ref{fig: result}j,~\ref{fig: result}k and \ref{fig: result}l, MIT* outperforms SOTA planners by quickly finding an initial solution and converging to the optimal one.}

\begin{table}[t]
\caption{Benchmarks evaluation comparison (unit seconds) (Fig.~\ref{fig: result})}
\centering
\resizebox{0.485\textwidth}{!}{
\begin{tabular}{lcccccccc}
 % \hline
 % \multicolumn{10}{|c|}{Contact Change Detection Accuracy} \\
 \toprule
 % \rowcolor{gray!10} % 

 &\multirow{2}*{$\textcolor{black}{\textit{$\mathcal{C}$-space}}$}
 &\multicolumn{3}{c}{${\text{EIT* (SOTA)}}$} 
 &\multicolumn{3}{c}{${\text{MIT* (ours)}}$} 
 &\multirow{2}*{\large$t^\textit{med}_\textit{init}\Uparrow$ (\%)}\\
\cmidrule(lr){3-5} \cmidrule(lr){6-8}
    % \textit{ENV} % 
    &\textcolor{black}{time~$(10^{-4})$}
    &$t^\textit{med}_\textit{init}$ &$c^\textit{med}_\textit{init}$ &$c^\textit{med}_\textit{final}$ &$t^\textit{med}_\textit{init}$ &$c^\textit{med}_\textit{init}$ &$c^\textit{med}_\textit{final}$ \\
    
 \toprule

    $\textnormal{FG}-\mathbb{R}^4$ &\textcolor{black}{1.9514}   &0.0068 &1.7640 &0.7350 &0.0053 &0.5227 &0.7158  &22.05\\
    $\textnormal{FG}-\mathbb{R}^8$ &\textcolor{black}{2.0918}  &0.0085  &2.7520  &1.7887 &0.0065 &{2.4283} &{1.6745} &{23.52}  \\
    $\textnormal{FG}-\mathbb{R}^{16}$ &\textcolor{black}{2.2267}   &0.0129   &4.4062   &3.1841 &{0.0083} &{4.1485} &{3.1171}  &{\textbf{35.65}}  \\
\midrule
    $\textnormal{RR}-\mathbb{R}^4$ &\textcolor{black}{1.6610}   &0.0423   &1.4779   &1.1062 &0.0303 &1.3375 &1.0807 &28.36  \\
    $\textnormal{RR}-\mathbb{R}^8$ &\textcolor{black}{1.6589}   &0.1671   &4.2433   &2.7571 &0.1213 &{3.9080} &2.7217  &{27.41}  \\
    $\textnormal{RR}-\mathbb{R}^{16}$ &\textcolor{black}{2.0987}   
    &{0.2094} &{4.6672} &{3.5319} &{0.1423} &{4.3738} &{3.3797}  &{32.04}\\
\midrule    
    $\textnormal{DW}-\mathbb{R}^4$ &\textcolor{black}{1.9932}    &0.0268 &2.3135 &1.3952 &0.0192 &2.0259 &1.3339  &28.73\\
    $\textnormal{DW}-\mathbb{R}^8$ &\textcolor{black}{2.0128}  &0.0350  &3.8081  &2.3013 &{0.0233} &{3.4246} &{2.2455} &{33.42}  \\
    $\textnormal{DW}-\mathbb{R}^{16}$ &\textcolor{black}{2.3486}   &0.0661   &6.2456   &3.8538 &{0.0389} &{5.7522} &{3.7675}  &{\textbf{41.15}}  \\
\midrule
    $\textnormal{GE}-\mathbb{R}^2$ &\textcolor{black}{1.0055}    &0.0065 &1.1004 &0.8688 &0.0051 &0.9743 &0.8685  &21.54\\
    $\textnormal{GE}-\mathbb{R}^4$ &\textcolor{black}{2.0734}  &0.0075  &1.3792  &0.9203 &{0.0052} &{1.2006} &{0.9036} &{30.67}  \\
    $\textnormal{GE}-\mathbb{R}^{8}$ &\textcolor{black}{2.2283}   &0.1203   &1.5726   &1.3775 &{0.0157} &{1.5686} &{1.3676}  &{\textbf{86.95}}  \\
\bottomrule
%  success & \multicolumn{3}{|c}{0.48} & \multicolumn{3}{|c}{0.71} & \multicolumn{1}{|c}{\textbf{0.88}} \\
%  % power  & & & & & &\\
%  \hline
\end{tabular}} \label{tab:benchmark}
\vspace{-1.7em} 
\end{table}
\begin{figure*}[t!]
    \centering
    \begin{tikzpicture}
    \node[inner sep=0pt] (russell) at (0,0)
    {\includegraphics[width=0.98\textwidth]{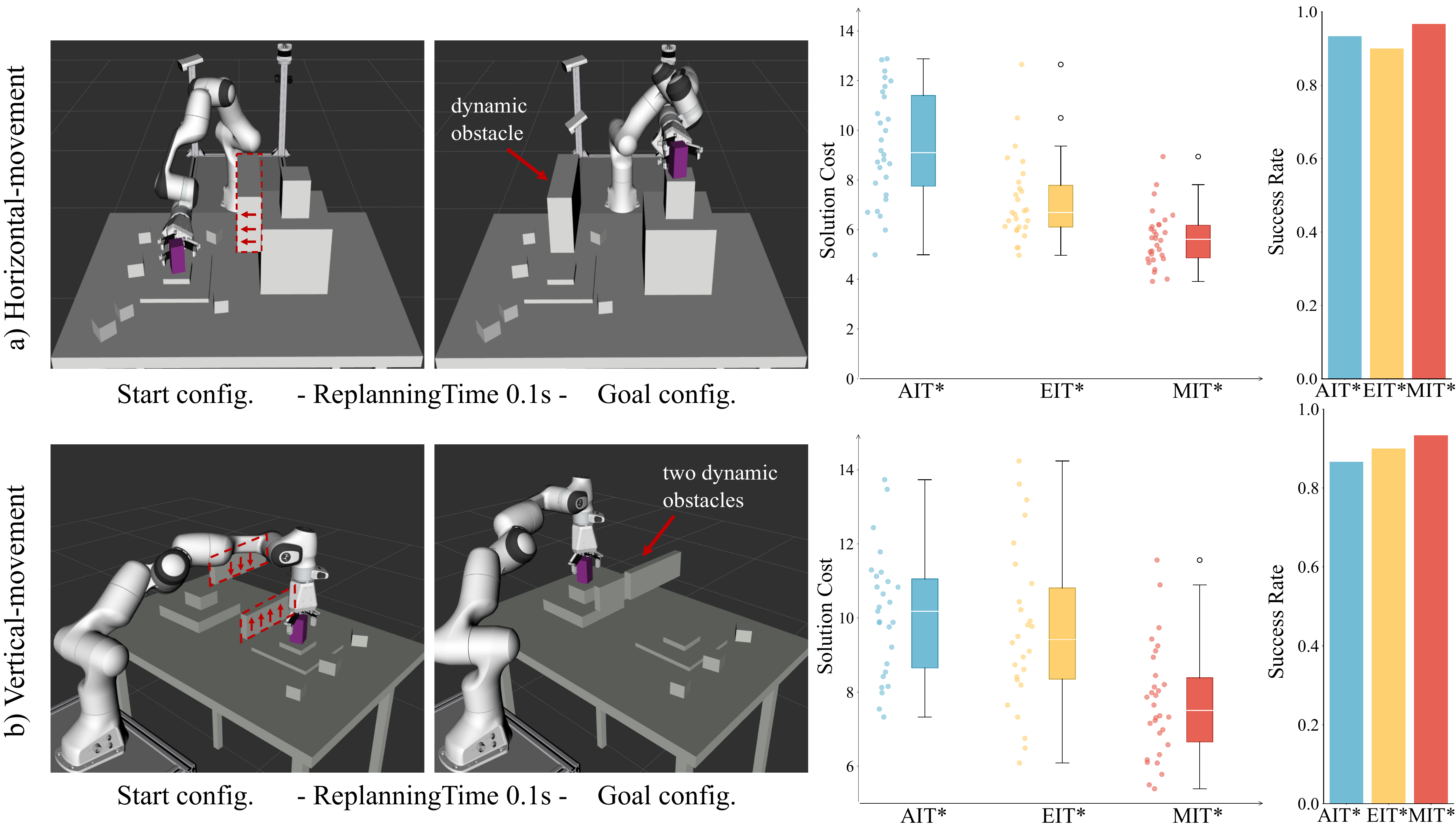}};

    \end{tikzpicture}
    \vspace{-0.3em} 
    \caption{\textcolor{black}{Detailed planner performance in dynamic experimental scenarios from Section~\ref{subsec:dynamicExpri} are summarized above. Fig.~\ref{fig: dynamicResult}a and b illustrates the robotic arm's start and goal configurations, along with the motion directions (i.e., horizontal and vertical) of the dynamic obstacles (red arrows). The robotic arm repeatedly travels between the start and goal configurations within the scenes, performing real-time re-planning (MaxTime 0.1s) to evaluate its dynamic obstacle avoidance capabilities (with unsuccessful runs assigned an infinite cost).}}
    \label{fig: dynamicResult}
    \vspace{-1.7em}
\end{figure*}
\textcolor{black}{As shown in Table~\ref{tab:benchmark}, MIT* consistently outperforms SOTA in initial median time ($t^\textit{med}_\textit{init}$) across various scenarios. For instance, in the $\textnormal{FG}-\mathbb{R}^{16}$ scenario, MIT* achieves a $t^\textit{med}_\textit{init}$ of 0.0083s, which is 35.65\% faster than EIT*. Similarly, in the $\textnormal{RR}-\mathbb{R}^{16}$ scenario, MIT* reduces $t^\textit{med}_\textit{init}$ by 32.04\%.
In more confined settings, such as $\textnormal{DW}-\mathbb{R}^{16}$ and $\textnormal{GE}-\mathbb{R}^{8}$, the level of improvement is even more substantial, with MIT* achieving a 41.15\% and 86.95\% reduction, respectively.}

Overall, the table highlights the superior efficiency of MIT* in reducing initial median times and enhancing the effectiveness of path planning algorithms.
\vspace{-1em}
\subsection{Real-world Path Planning Tasks}\label{subsec:realExpri}
\textcolor{black}{We compare MIT* with SOTA optimal path planners via the base-manipulator robot (DARKO) to evaluate performance in converging to the optimal solution and success rate over 30 runs (Fig.~\ref{fig: Realresult}). The \textit{Beer Barrel}-ENV includes simple cup holder obstacles, while the \textit{Shelf}-ENV and \textit{Kitchen}-ENV feature navigating through cluttered, narrow spaces. Each scenario requires finding a collision-free path from start to goal.}
\subsubsection{\textbf{Beer barrel cup placement task}}
Fig.~\ref{fig: Realresult}a illustrates the start and goal configurations for the cup placement task. In this scenario, a single robotic manipulator is used to grab a beer cup and position it under the beer tap of a keg, all while avoiding obstacles.
All planners were given 1.0 seconds to address the beer barrel cup placement problem. Over the course of 30 trials, MIT* achieved a 93.33\% success rate with a median solution cost of 13.2741. EIT* had a success rate of 76.67\% with a median solution cost of 16.8917. AIT* was 63.33\% successful, with a median solution cost of 18.6277.
\subsubsection{\textbf{Industry shelf container rearrangement task}}
The initial and final configurations for the shelf task are shown in Fig.~\ref{fig: Realresult}b. The robot grasps an industry-standard (\textit{tolerance} $\leq$ 5mm) container from a lower position and positions it on the upper layer between two larger containers. The challenge lies in precisely inserting the container into narrow spaces, making the planning of a collision-free path particularly difficult.
Each planner was allocated 1.5 seconds to solve the constrained lift-up and insertion problem within a limited space. Across 30 trials, MIT* achieved an 80\% success rate with a median solution cost of 9.7145. EIT* had a 60\% success rate with a median solution cost of 11.1054. AIT* managed a 53.33\% success rate with a median solution cost of 12.7851.
\subsubsection{\textbf{Kitchen model pan cooking task}}
The DARKO robot was positioned in front of a kitchen model for the third task. The start and goal configurations are shown in Fig.~\ref{fig: Realresult}c. This task is challenging because the manipulator must maneuver the pan within a cluttered oven while avoiding collisions with both the base robot and the kitchen shelves.
Each planner was given 5.0 seconds to solve the kitchen pan reallocation problem. Over 30 trials, MIT* achieved a 36.67\% success rate with a median solution cost of 15.9380. EIT* had a success rate of 26.67\% and a median solution cost of 21.2667. AIT* recorded a 20\% success rate with a median solution cost of 23.5611.

In short, compared with the AIT* and the EIT*, the MIT* achieves the best performance in finding the initial solution and converging to the optimal solution.
\textcolor{black}{\subsection{Dynamic Experimental Re-planning Tasks}\label{subsec:dynamicExpri}
We evaluate MIT* with SOTA optimal path planners using the robot arm to benchmark its performance in terms of the optimal solution costs and success rate over 30 runs (Fig.~\ref{fig: dynamicResult}). The \textit{Horizontal-movement}-ENV scenario involves manipulating blocks while avoiding a single dynamic obstacle, whereas the \textit{Vertical-movement}-ENV scenario features navigation through two vertically moving dynamic obstacles traveling in opposite directions. Each scenario requires computing a collision-free path from the start to the goal configuration.
\subsubsection{\textbf{Horizontal-movement block task}}
Fig.~\ref{fig: dynamicResult}a illustrates the start and goal configurations for the horizontal movable block task. In this scenario, a single robotic manipulator is used to grab a table block and navigate between the start and goal positions five times, all while avoiding obstacles.
All planners were allocated 0.1 seconds to solve the problem. Across 30 trials, for forward planning from the start to the goal configuration, AIT* achieved an average cost of 8.8167 with a 93.33\% success rate, while EIT* achieved 6.7306 with a 90\% success rate. Our proposed MIT* attained an average cost of 5.5329 with a 96.67\% success rate, representing a 17.78\% cost improvement over the best-performing baseline planner.
\subsubsection{\textbf{Two vertical-movement blocks task}}
As illustrated in Fig.~\ref{fig: dynamicResult}b, the task starts from the right part of the table and ends at the left. Two dynamic obstacles move repeatedly, one from top to bottom and the other from bottom to top. The robot arm needs to pick up a block from the start and place it at the goal, repeating this process five times. Avoiding collisions with two opposite moving obstacles, constrained narrow space, and limited planning time make the planning problem challenging.
Each planner was allocated 0.1 seconds of re-planning time to solve the back-and-forth problem within a limited space. Across 30 trials, AIT* achieved an 86.67\% success rate with a median solution cost of 10.2138, while EIT* reached a 90\% success rate with a median cost of 9.6934. Our proposed MIT* attained a 93.33\% success rate and a median cost of 7.8718, resulting overall in an 18.79\% cost improvement.
}

\textcolor{black}{In summary, MIT* quickly re-planning and converges to high-quality paths, it effectively supports robotic task execution in constrained and dynamic environments.}
\subsection{Discussion}
\subsubsection{Comparison with SOTA Planners}To evaluate the performance of MIT*, we compared it with SOTA planners using success rate and solution cost metrics across three real-world and nine simulation tasks. MIT* demonstrated superior performance
in all scenarios.
In real-world tasks, MIT* achieved notable improvements: in the \textit{Beer Barrel}-ENV, it reduced solution cost by 21.55\% and improved success rate by 16.67\% over EIT*. Compared to AIT*, MIT* lowered cost by 28.73\% and raised success rate by 30\%. In the \textit{Shelf}-ENV, MIT* reduced cost by 12.55\% and improved success rate by 20\% compared to EIT*, and reduced cost by 24.02\% with a 26.67\% higher success rate than AIT*. In the \textit{Kitchen}-ENV, MIT* outperformed EIT* by reducing cost by 25\% and increasing success rate by 10\%, and compared to AIT*, MIT* lowered cost by 32.74\% while increasing success rate by 16.67\%. 
%
% Overall, MIT* integrates estimated informed sets for fast initial path finding, an adaptive sampler for dynamic settings, and length-related collision checks for reverse search, achieving the highest success rates with the shortest paths.

\begin{figure}[t!]
    \centering
    \begin{tikzpicture}
    %first
    \node[inner sep=0pt] (russell) at (-4.2,0)
    {\includegraphics[width=0.23\textwidth]{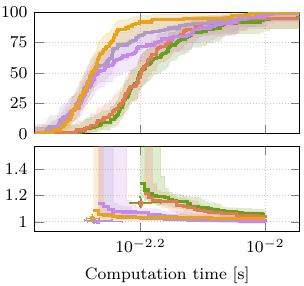}};
    \node[inner sep=0pt] (russell) at (0,0)
    {\includegraphics[width=0.23\textwidth]{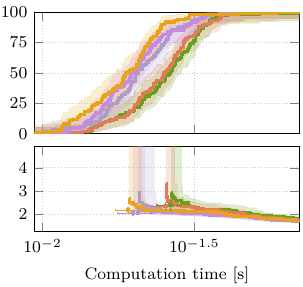}};  

    \node[inner sep=0pt] (russell) at (-6.5,0.35){\includegraphics[width=0.023\textwidth]{figure/benchmark/secc_cost.pdf}};
    
    \node[inner sep=0pt] (russell) at (-2.1,-2.3){\includegraphics[width=0.45\textwidth]{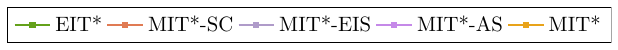}};
    
    \node at (-4.2,2.2) {\footnotesize (a) GE in $\mathbb{R}^{2}$ - MaxTime: 0.01s};
    \node at (0.2,2.2) {\footnotesize (b) DW in $\mathbb{R}^{4}$ - MaxTime: 0.07s};

    \end{tikzpicture}
    \vspace{-1.4em} 
    \caption{Detailed ablation study results from Section~\ref{subsec:ablation} are summarized.  Cost plots show solution cost and time, with lines for optimal cost progression and error bars for 99\% confidence intervals.
}
    \label{fig: ablaResult}
    \vspace{-1.7em}
\end{figure}

\textcolor{black}{Through simulations and real-world experiments, MIT* is shown to explore and converge more efficiently under the guidance of EIS and an adaptive sampler for dynamic settings, enabling the rapid discovery of feasible solutions.}
\textcolor{black}{\subsubsection{Ablation Study for Contributions}\label{subsec:ablation} To evaluate the individual contributions of the proposed planner, we performed an ablation study focusing on the adaptive sampler (MIT*-AS), estimated informed set (MIT*-EIS), and length-related sparse collision checking (MIT*-SC) in two representative scenarios (Fig.~\ref{fig: ablaResult}): goal enclosure (GE)-$\mathbb{R}^{2}$ and dividing walls (DW)-$\mathbb{R}^{4}$. In the GE scenario, which lacks narrow passages, the adaptive sampler was less effective as the bridge sampling component was underutilized, resulting in inferior performance compared to MIT*-EIS, which could initially prune the search space. Sparse collision checking offered limited improvement here, slightly reducing path cost but increasing initial solution time due to additional collision checks. In contrast, in the DW scenario, which features numerous narrow passages, the adaptive sampler demonstrated certain advantages, as both the bridge test and Gaussian sampling were effective, outperforming MIT*-EIS. Sparse collision checking showed benefits in this scenario by improving reverse search precision and reducing the need for costly restarts. Overall, MIT*, integrating all contributions, delivered better performance, highlighting the synergistic nature of the proposed methods.}
\subsubsection{Limitations and Future Work} While MIT* demonstrates superior performance, it also has some limitations. The proposed EIS may be overly conservative in environments with irregular cost landscapes or poorly aligned heuristics. In such cases, the repeated enlargement of the EIS could introduce computational overhead, potentially negating the method's advantages. Although we utilized a dynamic adjustment mechanism to iteratively refine the EIS, its effectiveness relies on the complexity of the problem and the quality of the initial solution cost. Furthermore, the current objective of our approach is to optimize path costs in Euclidean space. This limits the EIS's applicability to other cost metrics and non-Euclidean spaces, where the EIS may not effectively constrain the search space. Expanding the EIS to include broader optimal objectives (e.g., obstacle clearance, energy efficiency, etc.) and dynamic scenarios (i.e., via local motion planners) is crucial for future works to enhance robustness. \textcolor{black}{In future work, we will integrate our planner with an RGB-D/LiDAR perception stack that continuously refreshes an occupancy or signed-distance map, enabling closed-loop planning around non-convex and dynamic obstacles.}

\section{Conclusion}\label{sec:conclu}
In this article, we proposed Multi-Informed Trees (MIT*), a sampling-based planner that is multi-informed by prior admissible cost (i.e., estimated informed set) and current solution cost (i.e., informed set). The novel estimated informed set can improve early-stage exploration, the adaptive sampling strategy can refine distribution in critical regions, and length-related adaptive sparse collision checks for edges can optimize lazy reverse search. Then, the probabilistic completeness and asymptotic optimality were guaranteed. Through simulation across dimensions and real-world experiments, we demonstrated that our algorithm shows faster initial path convergence and shorter path length over the SOTA algorithms in a robust manner. Moreover, real-world tasks can further demonstrate the effectiveness of our method in practical applications.  

% Future research directions could be integrating human awareness and safety constraints into MIT* for human-interaction planning, and using learning-based techniques to enhance sampling and motion planning performance.

% References

\bibliographystyle{IEEEtran}
\bibliography{reference.bib}

% \setstretch{0.92}
\begin{IEEEbiography}[{\includegraphics[width=1in,height=1.25in,clip,keepaspectratio]{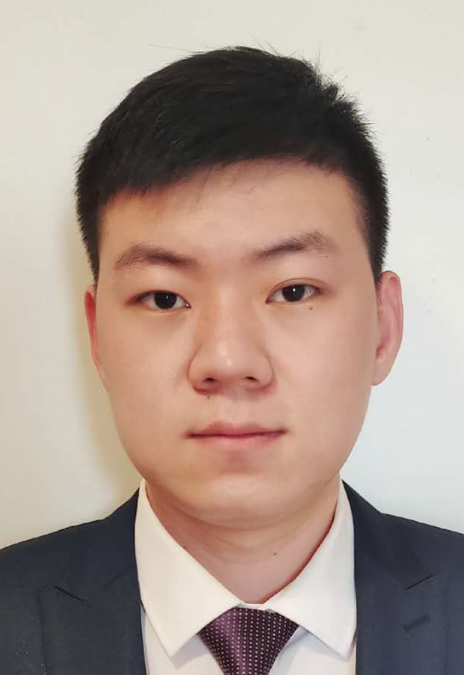}}] {Liding Zhang} is currently a Ph.D. candidate at the School of Computation, Information and Technology (CIT) chair of informatics 6, Technical University of Munich, Germany. He received the B.Sc. degree in mechanical engineering from the Rhine-Waal University of Applied Sciences, Germany, in 2020 and the M.Sc. degree in mechanical engineering and automation from the Technical University of Clausthal, Germany, in 2022.\\
\indent His current research interests include robotic task and motion planning, multi-robot collaborations.
\end{IEEEbiography}

% \vspace{30pt}

\begin{IEEEbiography}[{\includegraphics[width=1in,height=1.25in, clip,keepaspectratio]{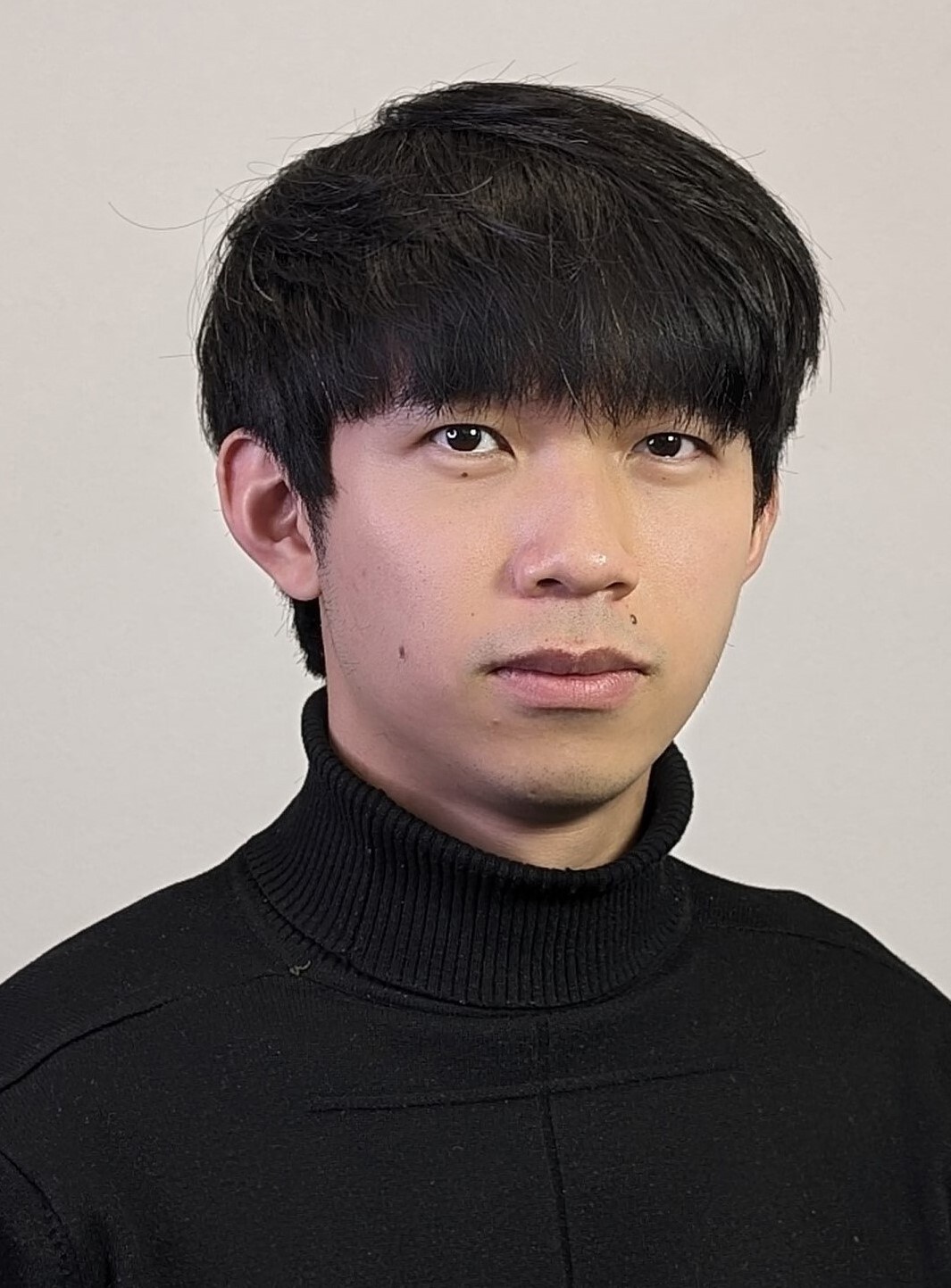}}] {Kuanqi Cai} is currently a research associate at the Munich Institute of Robotics and Machine Intelligence (MIRMI), Technical University of Munich. He worked as a research assistant at The Chinese University of Hong Kong, Hong Kong, from 2019 to 2020. Following that, he served as a visiting student researcher at the Southern University of Science and Technology. In 2021, he was a Robotics Student Fellow at ETH Zurich. In 2022, he was employed at The Chinese University of Hong Kong (Shenzhen Research Institute). He obtained his B.E. degree from Hainan University in 2018 and his M.E. degree from Harbin Institute of Technology in 2021. His current research interests include motion planning and human-robot interaction.
\end{IEEEbiography}

% \vspace{30pt}

\begin{IEEEbiography}[{\includegraphics[width=1in, height=1.25in, clip, keepaspectratio]{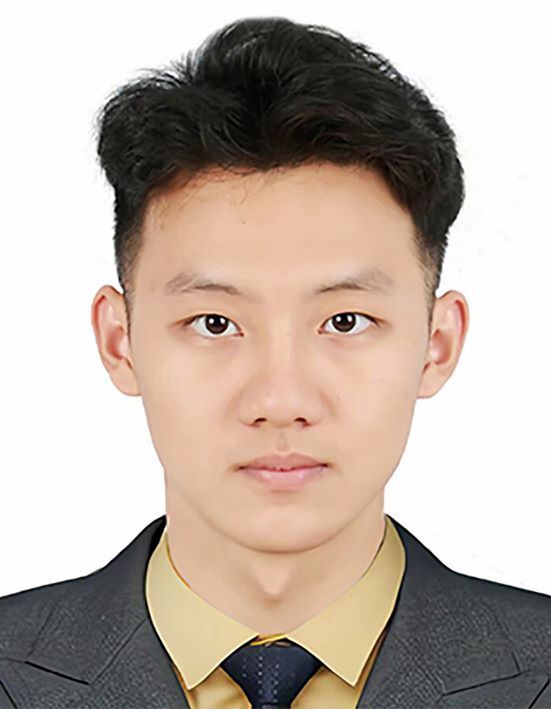}}]{Yu Zhang} received the M.Eng. degree from the School of Intelligence Science and Technology, University of Science and Technology Beijing, Beijing, China, in 2022. He is currently working toward the Ph.D. degree in computer science as a member of the Informatics 6, Technical University of Munich, Munich, Germany.\\
\indent His current research interests include optimization and control in robotics, machine learning, adaptive and learning control.
\end{IEEEbiography}

% \vspace{40pt}

\begin{IEEEbiography}[{\includegraphics[width=1in,height=1.25in,clip,keepaspectratio]{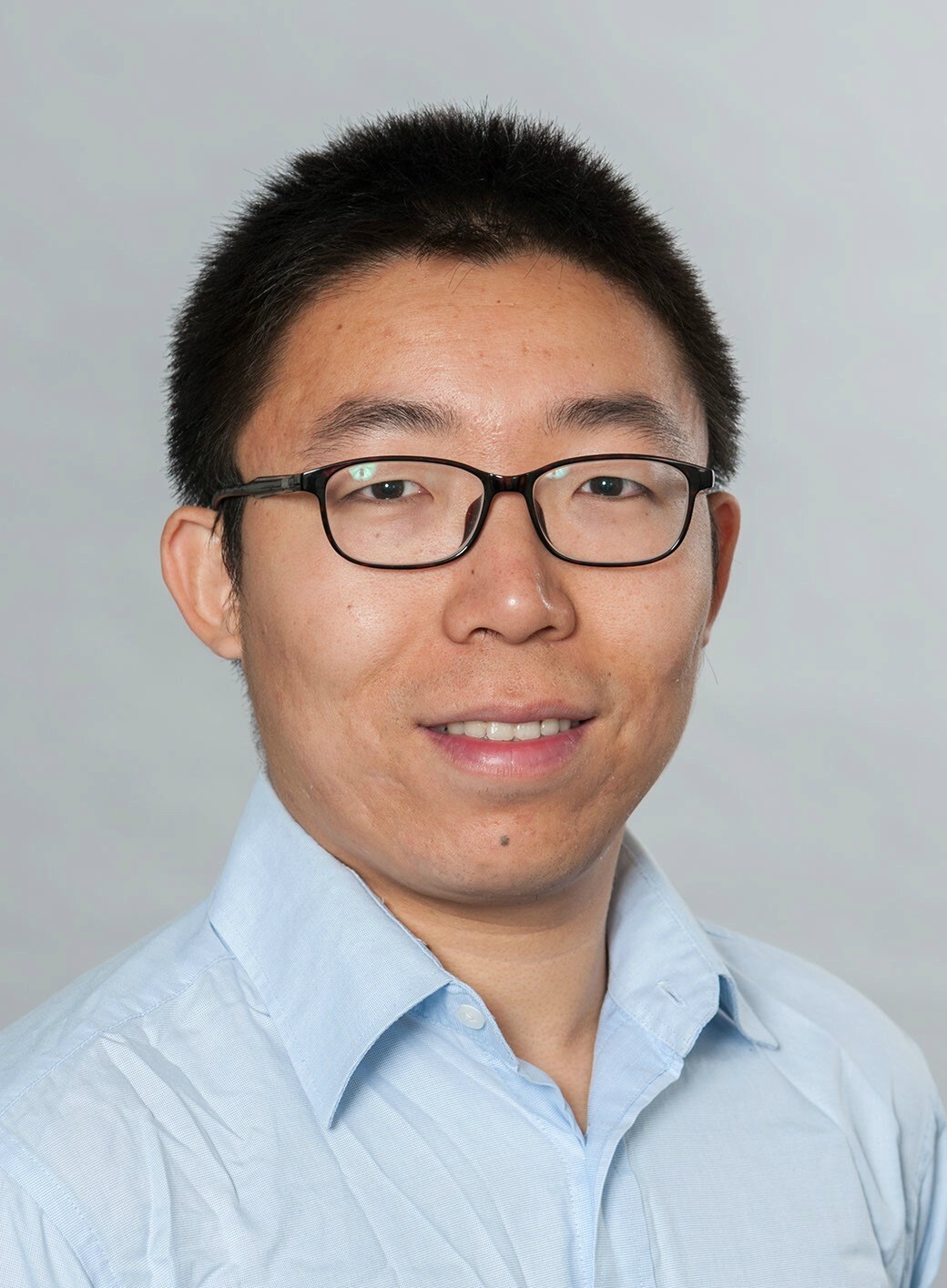}}] {Zhenshan Bing} (Member, IEEE) received the B.S. degree in mechanical design, manufacturing, and
automation and the M.Eng. degree in mechanical
engineering from Harbin Institute of Technology,
China, in 2013 and 2015, respectively, and the
Ph.D. degree in computer science from the Technical University
University of Munich, Germany, in 2019.\\
\indent From 2019 to 2024, he was a Post-Doctoral
Researcher with the Department of Informatics,
Technical University of Munich. He is currently an
Associate Professor with the School of Intelligence
Science and Technology, Nanjing University, Suzhou. His research interests
include bio-inspired robotics and embodied intelligence control algorithms.
\end{IEEEbiography}

% \vspace{30pt}

\begin{IEEEbiography}[{\includegraphics[width=1in,height=1.25in,clip,keepaspectratio]{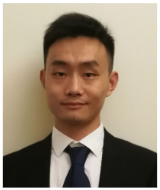}}]{Chaoqun Wang} (Member, IEEE) received the B.E. degree in automation from Shandong University, Jinan, China, in 2014, and the Ph.D. degree in robot and artificial intelligence from the Department of Electronic Engineering, The
Chinese University of Hong Kong, Hong Kong, in 2019.
During his Ph.D. study, he spent six months with the University of British Columbia, Vancouver, BC, Canada, as a Visiting Scholar. He was a Postdoctoral Fellow with the Department of Electronic Engineering, The Chinese University of Hong Kong, from 2019 to 2020. He is currently a Professor with the School of Control Science and Engineering, Shandong University. His current research interests include autonomous vehicles, active and autonomous exploration, and path planning.
\end{IEEEbiography}

% \vspace{-30pt}

\begin{IEEEbiography}[{\includegraphics[width=1in,height=1.25in,clip,keepaspectratio]{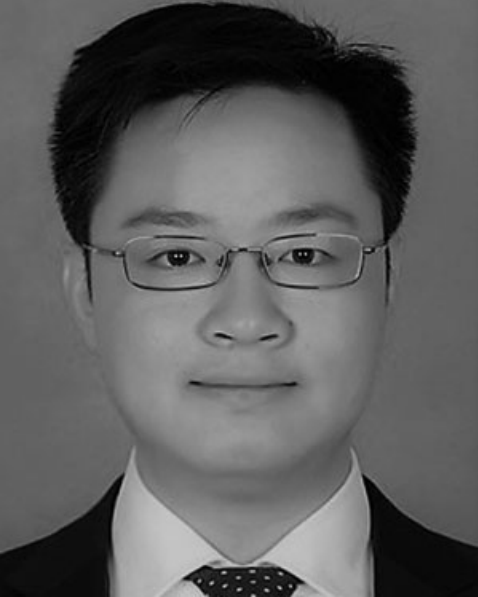}}] {Fan Wu} (Member, IEEE) received the B.Sc. degree
in mathematics from the University of Science and
Technology Beijing, Beijing, China, in 2012, and the
M.Sc. degree in financial mathematics from King's
College London, London, U.K. Since 2016, he has
been working toward the Ph.D. degree in robotics
from King's College London, U.K. He is currently a postdoctoral researcher with the Munich Institute of Robotics and Machine Intelligence (MIRMI), Technical University of Munich, Germany.
His research interests span the topics in compliant
robotics, optimal control and reinforcement learning.

\end{IEEEbiography}

% \vspace{-30pt}

\begin{IEEEbiography}[{\includegraphics[width=1in,height=1.25in,clip,keepaspectratio]{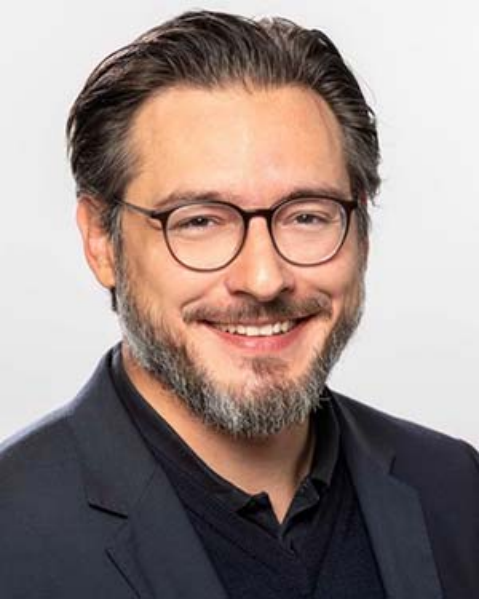}}]{Sami Haddadin} (IEEE Fellow) received the
Dipl.-Ing. degree in electrical engineering in 2005,
the M.Sc. degree in computer science in 2009 from
the Technical University of Munich (TUM), Munich,
Germany, the Honours degree in technology management in 2007 from Ludwig Maximilian University,
Munich, Germany, and TUM, and the Ph.D. degree
in safety in robotics from RWTH Aachen University,
Aachen, Germany, in 2011. He is currently a Full
Professor and the Chair with Robotics and Systems
Intelligence, TUM, and the Founding Director of the
Munich Institute of Robotics and Machine Intelligence (MIRMI), Munich.
Dr. Haddadin was the recipient of numerous awards for his scientific work,
including the George Giralt Ph.D. Award (2012), IEEE RAS Early Career Award
(2015), the German President's Award for Innovation in Science and Technology
(2017), and the Leibniz Prize (2019).
\end{IEEEbiography}

% \vspace{-30pt}

\begin{IEEEbiography}[{\includegraphics[width=1in,height=1.25in,clip,keepaspectratio]{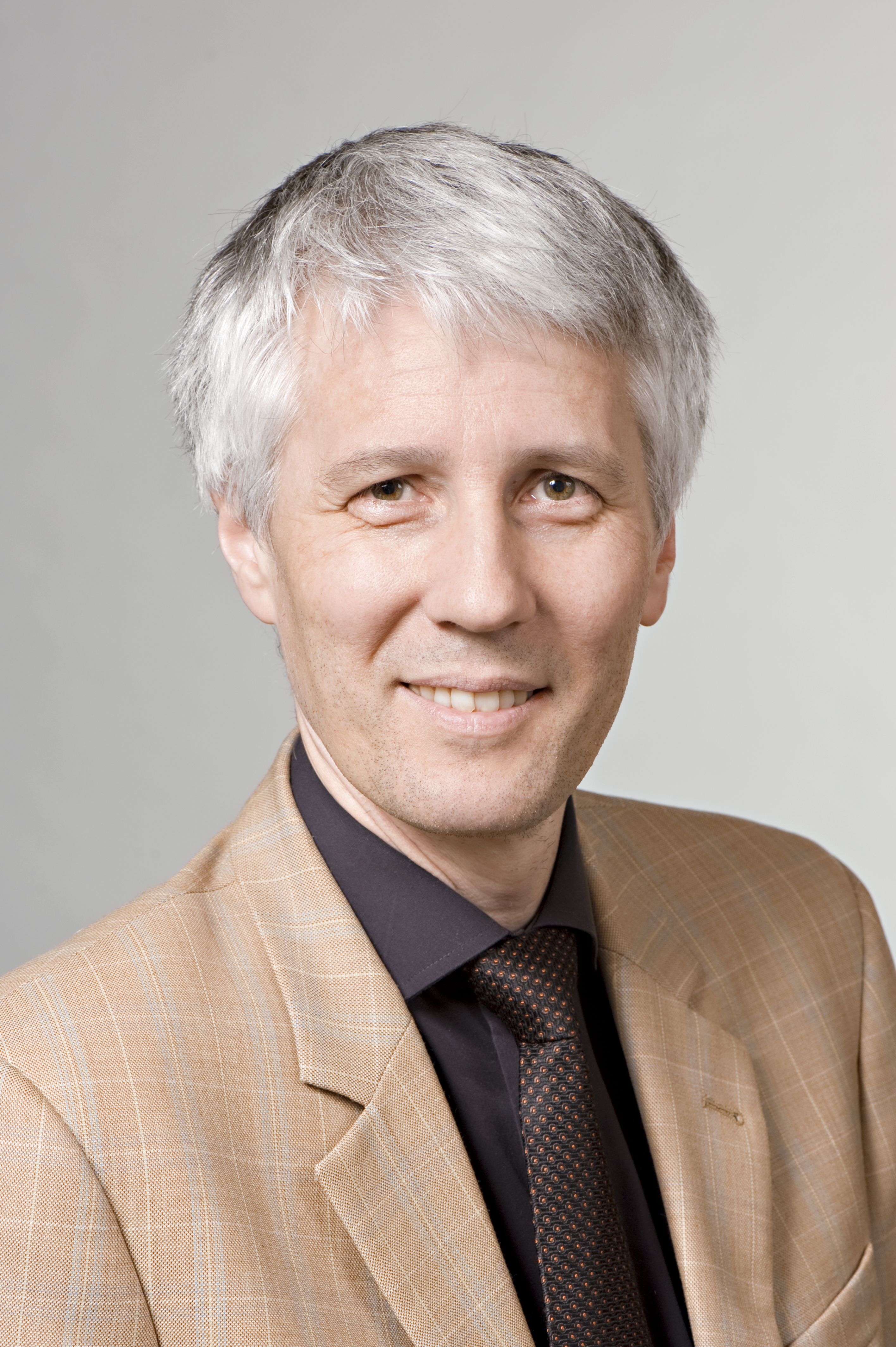}}]{Alois Knoll} (IEEE Fellow) received his diploma
(M.Sc.) degree in Electrical/Communications
Engineering from the University of Stuttgart,
Germany, in 1985 and his Ph.D. (summa cum
laude) in Computer Science from Technical University of Berlin, Germany, in 1988. He served on
the faculty of the Computer Science department
at TU Berlin until 1993. He joined the University
of Bielefeld, Germany as a full professor and
served as the director of the Technical Informatics research group until 2001. Since 2001,
he has been a professor at the Department of Informatics, Technical
University of Munich (TUM), Germany . He was also on the board of
directors of the Central Institute of Medical Technology at TUM (IMETUM). 
% From 2004 to 2006, he was Executive Director of the Institute of Computer Science at TUM. Between 2007 and 2009, he was a member of the EU's highest advisory board on information technology, ISTAG, the Information Society Technology Advisory Group, and a member of its subgroup on Future and Emerging Technologies (FET). In this capacity, he was actively involved in developing the concept of the EU's FET Flagship projects. 
His research interests include cognitive, medical
and sensor-based robotics, multi-agent systems, data fusion, adaptive
systems, multimedia information retrieval, model-driven development of
embedded systems with applications to automotive software and electric
transportation, as well as simulation systems for robotics and traffic.

\end{IEEEbiography}

% \vspace{-50pt}

\vfill

\end{document}